\documentclass{article}

\usepackage[preprint]{neurips_2026}

\usepackage[utf8]{inputenc} 
\usepackage[T1]{fontenc}    
\usepackage{hyperref}       
\usepackage{url}            
\usepackage{booktabs}       
\usepackage{amsmath}        
\usepackage{amsfonts}       
\usepackage{nicefrac}       
\usepackage{microtype}      
\usepackage[table]{xcolor}  
\usepackage{makecell}       
\usepackage{multirow}       
\usepackage{graphicx}
\usepackage{tcolorbox}
\usepackage{wrapfig}
\usepackage{subcaption}

\title{LLaVA-UHD v4: What Makes Efficient Visual Encoding in MLLMs?}

\author{
  Kechen Fang\textsuperscript{1} \quad
  Yihua Qin\textsuperscript{1} \quad
  Chongyi Wang\textsuperscript{2} \quad
  Wenshuo Ma\textsuperscript{2} \quad
  Tianyu Yu\textsuperscript{1} \quad
  Yuan Yao\textsuperscript{1}\thanks{Corresponding author}\\
  \textsuperscript{1}Tsinghua University
  \textsuperscript{2}ModelBest
}

\begin{document}

\maketitle

\begin{abstract}

Visual encoding constitutes a major computational bottleneck in Multimodal Large Language Models (MLLMs), especially for high-resolution image inputs. The prevailing practice typically adopts global encoding followed by post-ViT compression. Global encoding produces massive token sequences, while post-ViT compression incurs the full quadratic attention cost of the ViT before any token reduction takes place.
In this work, we revisit this convention along two dimensions: the encoding strategy and visual token compression.
First, controlled experiments show that slice-based encoding outperforms global encoding across benchmarks, suggesting that preserving local details through sliced views can be more beneficial than applying global attention for fine-grained perception. 
Second, we introduce intra-ViT early compression, which reduces tokens in shallow ViT layers and substantially lowers visual-encoding FLOPs while preserving downstream performance.
By integrating intra-ViT compression into the slice-based encoding framework, we present LLaVA-UHD v4, an efficient and compute-controllable visual encoding scheme tailored for high-resolution inputs. Across a diverse set of benchmarks covering document understanding, OCR, and general VQA, LLaVA-UHD v4 reduces visual-encoding FLOPs by $55.8\%$ while matching or even surpassing baseline performance. These results suggest that visual-encoding efficiency can be substantially improved without sacrificing downstream performance, providing a practical design direction for efficient high-resolution MLLMs. All model weights and code will be publicly released to support further research\footnote[1]{Code available at \url{https://github.com/THUMAI-Lab/LLaVA-UHD-v4}}.
\end{abstract}

\section{Introduction}

Multimodal Large Language Models (MLLMs) have made remarkable progress on a broad spectrum of vision-language tasks~\cite{liu2023visual, li2023blip, yao2024minicpm, bai2023qwenvlversatilevisionlanguagemodel}. As the field shifts toward fine-grained perception~\cite{Mathew_2021_WACV, ouyang2025omnidocbench, masry2022chartqa} and detailed image understanding~\cite{zhang2024mme, wu2024v}, high-resolution image inputs are rapidly becoming the default. To preserve as much visual detail as possible and sustain downstream performance, the prevailing recipe is global encoding~\cite{wang2024qwen2, team2026kimi}, which feeds the full image directly into the vision encoder. As resolution grows, this yields a token sequence that scales with image area. To relieve the downstream LLM from this token explosion, mainstream frameworks then attach a compression module after the vision encoder~\cite{yao2024minicpm}. That is, visual tokens are reduced only after the vision encoder has already executed full global self-attention at quadratic complexity. This approach is straightforward to implement, yet its computational cost increases rapidly with resolution. Furthermore, post-ViT compression cannot mitigate the ViT's cost, as it only operates after the full computation has already occurred. This cost is far from negligible in the high-resolution regime, making high-resolution visual encoding a central efficiency bottleneck in modern MLLMs.

In this work, we systematically rethink this inefficient convention, beginning with the encoding paradigm. The community has widely held that global encoding is the more direct and lossless choice, since it supplies complete global context and allows arbitrary patch-to-patch interaction~\cite{wang2024qwen2, team2026kimi}. However, our empirical evaluations across diverse benchmarks yield a surprising conclusion that slice-based encoding consistently outperforms global encoding, suggesting that slice-based strategies can already provide sufficiently informative feature representations. Moreover, by processing large images via partitioning, slice-based encoding structurally sidesteps the quadratic blow-up incurred by global encoding, making it the more efficient paradigm for ultra-high-resolution images.

While slice-based encoding alleviates the per-forward attention explosion to some extent, high resolution still inherently produces a large number of tokens. Existing compression schemes, such as MLP-based spatial merging~\cite{wang2024qwen2, lu2025internvl}, Pixel-Shuffle and various resamplers ~\cite{li2023blip, alayrac2022flamingo} and token-pruning approaches ~\cite{bolya2022token}, are almost exclusively post-ViT. They only ease the burden on the downstream LLM and do nothing about the heavy cost inside the encoder itself. To achieve truly extreme efficiency, we must strike at the root of the bottleneck: the ViT's own compute. Intuitively, token compression must be moved inside the vision encoder and triggered as early as possible, so that the vast majority of ViT layers operate on only a small number of tokens. The vision encoder is typically a pretrained model, and inserting a randomly initialized compressor into its intermediate layers can perturb or even destroy its learned visual representations. Such modifications incur substantial additional training cost and offer no guarantee of recovering the original performance, making early in-ViT token compression a problem that demands careful design.

To address the challenges above, we introduce a parameter-reuse early compressor: a window-attention block coupled with a downsampling MLP, both inserted into the shallow layers of the ViT and initialized by reusing the pretrained weights of their adjacent ViT layers. This warm start places the new module very close to the representation manifold of the original ViT from the very first training step, thereby avoiding any disruption to the learned visual representations. The module compresses the ViT's tokens by $4\times$ at a very early stage of the encoder, so that the vast majority of subsequent ViT layers operate on only a small fraction of the original token budget.

Combining slice-based encoding with the proposed intra-ViT early compression, we obtain LLaVA-UHD v4, an efficient and compute-controllable visual encoding architecture for high-resolution MLLMs. 
Across eight standard benchmarks, LLaVA-UHD v4 matches or surpasses a post-ViT baseline at the same $16\times$ compression ratio in overall downstream accuracy.

Our main contributions are as follows: (1) We revisit the common practice of global encoding  and demonstrate the advantages of slice-based encoding in preserving fine-grained details while circumventing the quadratic computational overhead. (2) Building on this insight, we identify the limitations of post-ViT token compression and propose a novel intra-ViT shallow-layer compression architecture that directly addresses the computational bottleneck of visual encoding. (3) Integrating these two designs, we propose LLaVA-UHD v4, which combines slice-based encoding with an early compressor and maintains competitive performance while achieving a $55.75\%$ acceleration in visual encoding FLOPs.

\section{Rethinking High-Resolution Visual Encoding}
\label{sec:pilot_study}

We begin with a controlled study of two design choices that are central to high-resolution MLLMs: (1) How high-resolution images are encoded before entering the ViT. (2) How visual tokens are compressed along the pipeline. For both questions, we default to SigLIP 2~\cite{tschannen2025siglip} as the ViT backbone and Qwen3~\cite{yang2025qwen3} as the LLM, while fixing the training data and the total visual-token budget reaching the LLM, so that any observed difference is attributable solely to the dimension under study.

\subsection{Slice-based Encoding Outperforms Global Encoding}
\label{sec:encoding_paradigm}

The community has converged on global encoding (GE) as the actual choice for high-resolution MLLMs~\cite{wang2024qwen2, lu2025internvl}, on the intuitive grounds that feeding the full image to the ViT preserves complete global context and permits arbitrary patch-to-patch interaction. Slice-based encoding (SE)~\cite{guo2024llava, chen2024far}, which partitions the image into smaller views encoded independently, is typically framed as a computational compromise, which sacrifices global context for tractable per-forward cost. In this section we test this framing directly: under matched compression and training conditions, which paradigm actually delivers better downstream accuracy?

\textbf{Setup.} The two paradigms share the ViT backbone, projector, LLM, and the post-ViT compressor, differing only in how the image is presented to the ViT. GE rescales the image to at most $N \times 448^2$ pixels and processes it in a single forward pass. SE decomposes the image into a thumbnail and a set of slices laid out by an aspect-ratio-aware best-grid policy. We sweep two compression ratios ($4\times$, $16\times$) and two data scales (4M, 8M), and evaluate on the eight benchmarks. To comprehensively assess model performance, we conduct evaluations on a broad benchmark suite encompassing mathematics, OCR, and general VQA tasks.

\textbf{SE consistently outperforms GE, with larger gains at higher scales.}
Table~\ref{tab:encoding} reports the SigLIP-2-based comparison.
Across all four settings, SE outperforms GE on average, with gains ranging from $0.5$ to $1.7$ points.
The advantage also tends to increase with data scale, growing from $0.5$ to $1.2$ points under $4\times$ compression and from $0.5$ to $1.7$ points under $16\times$ compression.
In the SigLIP-2 sweep, the SE margin increases from 4M to 8M under both compression ratios, suggesting that the observed benefit persists with additional supervision in this setting.
In particular, the advantage is most pronounced on OCR-intensive tasks requiring fine-grained recognition, where SE leads GE by $3.6$ to $5.5$ points on OCRBench across the four SigLIP-2 settings.

\begin{table}[t!]
  \centering
  \caption{\textbf{Comparison of encoding strategies.} We compare the two encoding strategies under different compression rates and data scales using SigLIP 2 as the ViT backbone. GE denotes global encoding and SE denotes slice-based encoding.}
  \footnotesize
  \label{tab:encoding}
  \setlength{\tabcolsep}{4pt}
  \resizebox{\linewidth}{!}{
  \begin{tabular}{l|ccccccccc|c}
    \toprule
    Data Scale & Method & MMMU & MathVista & MMB$_\text{EN}$ & MMB$_\text{CN}$ & MMStar & HallBench & AI2D & OCRBench & Avg. \\
    \midrule
    \rowcolor{blue!7} \multicolumn{11}{c}{Compression Rate 4$\times$} \\
    \multirow{2}{*}{4M} & GE & 58.4 & \textbf{67.4} & \textbf{83.7} & \textbf{81.5} & \textbf{63.5} & 48.5 & 80.3 & 77.6 & 70.1 \\
     & SE & \textbf{61.9} & 66.7 & 82.9 & 79.5 & 62.3 & \textbf{49.1} & \textbf{80.5} & \textbf{82.0} & \textbf{70.6} \\
    \hline
    \multirow{2}{*}{8M} & GE & \textbf{60.4} & \textbf{71.4} & 84.4 & \textbf{83.5} & \textbf{65.4} & 49.3 & \textbf{82.5} & 80.0 & 72.1 \\
     & SE & 60.3 & 71.2 & \textbf{85.2} & 83.4 & 64.3 & \textbf{56.3} & 82.0 & \textbf{83.6} & \textbf{73.3} \\
    \rowcolor{blue!7} \multicolumn{11}{c}{Compression Rate 16$\times$} \\
    \multirow{2}{*}{4M} & GE & \textbf{58.4} & 62.7 & \textbf{80.3} & \textbf{81.9} & 60.4 & 47.7 & \textbf{78.5} & 72.0 & 67.7 \\
     & SE & 57.9 & \textbf{63.0} & 79.4 & 79.1 & \textbf{60.6} & \textbf{50.5} & 77.7 & \textbf{77.5} & \textbf{68.2} \\
    \hline
    \multirow{2}{*}{8M} & GE & \textbf{58.7} & 65.6 & 82.9 & \textbf{82.6} & 60.5 & 47.0 & \textbf{80.0} & 73.6 & 68.9 \\
     & SE & 58.6 & \textbf{67.3} & \textbf{83.7} & 82.3 & \textbf{62.9} & \textbf{51.2} & 79.8 & \textbf{79.1} & \textbf{70.6} \\
    \bottomrule
  \end{tabular}}
\end{table}

\begin{wraptable}{r}{0.48\textwidth}
  \vspace{-2mm}
  \centering
  \caption{\textbf{Robustness of slice-based encoding.} Average accuracy under (i) an alternative vision encoder backbone and (ii) a higher-resolution slicing schedule, both at compression rate $16\times$.}
  \footnotesize
  \setlength{\tabcolsep}{4pt}
  \renewcommand{\arraystretch}{1.1}
  \label{tab:encoding_robust}
  \begin{tabular}{c|c|c|c}
    \toprule
    Setting & Scale & GE & SE \\
    \midrule
    \multirow{2}{*}{MoonViT}
    & 8M  & 70.3 & \textbf{71.6} \\
    & 16M & 72.2 & \textbf{73.6} \\
    \midrule
    Higher-Res & 8M  & 68.8 & \textbf{71.0} \\
    \bottomrule
  \end{tabular}
  \vspace{-2mm}
\end{wraptable}

\textbf{Robustness.} To ensure that the observed advantage of SE is not attributable to a specific backbone or slicing configuration, we conduct two stress tests under more demanding conditions, with average accuracy reported in Table~\ref{tab:encoding_robust}. First, we replace SigLIP 2 with MoonViT~\cite{team2025kimi, team2026kimi}, a ViT explicitly pretrained on native-resolution inputs, where SE retains an average margin of approximately $+1.5$ points across both 8M and 16M data scales, indicating that its effectiveness generalizes across visual encoders. Second, under the $16\times$/8M setting, we adopt an alternative slicing schedule with a fourfold larger slice budget, which preserves higher per-image resolution and exposes the encoder to substantially more high-resolution visual tokens. Under this more demanding slicing configuration, the margin further widens to more than $+2$ points on average, with substantially larger gains on OCR-intensive tasks. Taken together, these results suggest that, under the resolution settings considered, the benefit of SE increases with input resolution and exhibits no evidence of saturation. Per-benchmark results for both stress tests are provided in Table~\ref{app:encoding_robust_details}.

\begin{tcolorbox}[colback=blue!3, colframe=blue!25, boxrule=0.5pt, arc=2pt]
\textit{\textbf{Finding 1.} Slice-based encoding consistently matches or outperforms global encoding across different compression rates, vision encoder backbones, and image resolutions.}

\end{tcolorbox}

\textbf{Analysis.} Across different backbones and slicing schedules, slice-based encoding (SE) consistently matches or outperforms global encoding (GE). We attribute this to a difference in inductive bias: SE preserves locality by decomposing the image into spatially coherent views, allowing the encoder to focus its capacity on fine-grained patterns within each slice, whereas GE processes the entire image jointly, forcing local details to compete with global context under a fixed token budget. A more detailed analysis is provided in Appendix~\ref{app:se_ge_analysis}.

\subsection{Compressing Visual Tokens at High Resolution}
\label{sec:compression}

Slice-based encoding (Section~\ref{sec:encoding_paradigm}) provides a stronger input pipeline, yet each high-resolution image still produces a large number of visual tokens that must be compressed before entering the LLM. These are conventionally compressed by a connector module placed between the ViT and the LLM. We address two questions about this scheme. First, which connector design performs best? Second, is this post-ViT compression sufficient enough at high resolution?

\begin{wraptable}{r}{0.48\textwidth}
  \vspace{-2mm}
  \centering
  \caption{\textbf{Connector comparison.}}
  \footnotesize
  \setlength{\tabcolsep}{4pt}
  \renewcommand{\arraystretch}{1.1}
  \label{tab:connector_avg}
  \begin{tabular}{c|c|c|c}
    \toprule
    Downsampling & Scale & Resampler & MLP \\
    \midrule
    \multirow{2}{*}{$4\times$}
    & 4M  & 65.51 & \textbf{69.10} \\
    & 8M  & 64.80 & \textbf{71.73} \\
    \midrule
    \multirow{3}{*}{$16\times$}
    & 4M  & 65.87 & \textbf{66.64} \\
    & 8M  & 67.66 & \textbf{68.84} \\
    & 16M & 70.39 & \textbf{70.81} \\
    \bottomrule
  \end{tabular}
  \vspace{-2mm}
\end{wraptable}

\textbf{Setup.} Two families dominate the connector designs. Query-based resamplers~\cite{bai2023qwenvlversatilevisionlanguagemodel, alayrac2022flamingo, li2023blip} attend a small set of learnable queries to the ViT output via cross-attention. Spatial-merging MLPs~\cite{liu2024improved, chen2024far} fold neighboring patch tokens via pixel-unshuffle and project them through a lightweight feed-forward network. We first compare both under matched conditions, sharing the ViT backbone, LLM, training recipe, slice-based encoding, and target token counts at $4\times$ and $16\times$ compression. Both are evaluated on the eight benchmarks of Section~\ref{sec:encoding_paradigm} across multiple data scales.

\textbf{MLP outperforms resampler.} Table~\ref{tab:connector_avg} reports the comparison results. The MLP connector outscores the resampler across all configurations, with the largest margins at lower compression ratios where it leads by $+3.3$ to $+6.7$ points at $4\times$.  We further observe that the gap narrows as the compression ratio tightens and training data scales up, falling to $+0.4$ points at $16\times$ compression with 16M training data, though MLP retains its lead in every cell.

\begin{tcolorbox}[colback=blue!3, colframe=blue!25, boxrule=0.5pt, arc=2pt]
\textit{\textbf{Finding 2.} Pixel-unshuffle-based MLP downsampling provides a stronger post-ViT compression baseline than query-based resampler.}
\end{tcolorbox}

\textbf{Analysis.} Pixel-unshuffle strictly preserves spatial structure by mapping each $k\times k$ ViT patch group into one token with concatenated channels, maintaining a coarse 2D layout. In contrast, the resampler uses content-agnostic learnable queries with global attention, discarding explicit spatial correspondence. The decisive factor is therefore not capacity (the resampler in fact uses more parameters at lower compression yet still loses by the largest margins) but whether spatial priors are built-in or must be learned. A more detailed analysis is provided in Appendix~\ref{app:connector_analysis}.

Together, \textit{Findings 1 and 2} establish slice-based encoding combined with an MLP connector as an effective baseline. However, because this token reduction occurs only after the vision encoder, it merely relieves the downstream LLM while leaving the ViT's massive internal compute entirely unchanged. To overcome this structural bottleneck, compression must be shifted inside the ViT pipeline. We detail the structure of our proposed intra-ViT compressor in Section~\ref{sec:method}.

\section{LLaVA-UHD v4}
\label{sec:method}

In this section, we answer the design questions raised at the end of Section~\ref{sec:compression} and introduce LLaVA-UHD v4. It builds on the slice-based encoding and MLP connector established in Section~\ref{sec:pilot_study} and adds an intra-ViT early compressor $\mathcal{D}$. We describe the end-to-end architecture in Section~\ref{sec:method_overview}, and introduce the design principles, structure, and parameter-reuse initialization in Section~\ref{sec:early_compression}.

\subsection{Overview}
\label{sec:method_overview}

\begin{figure}
    \centering
    \includegraphics[width=1.0\linewidth]{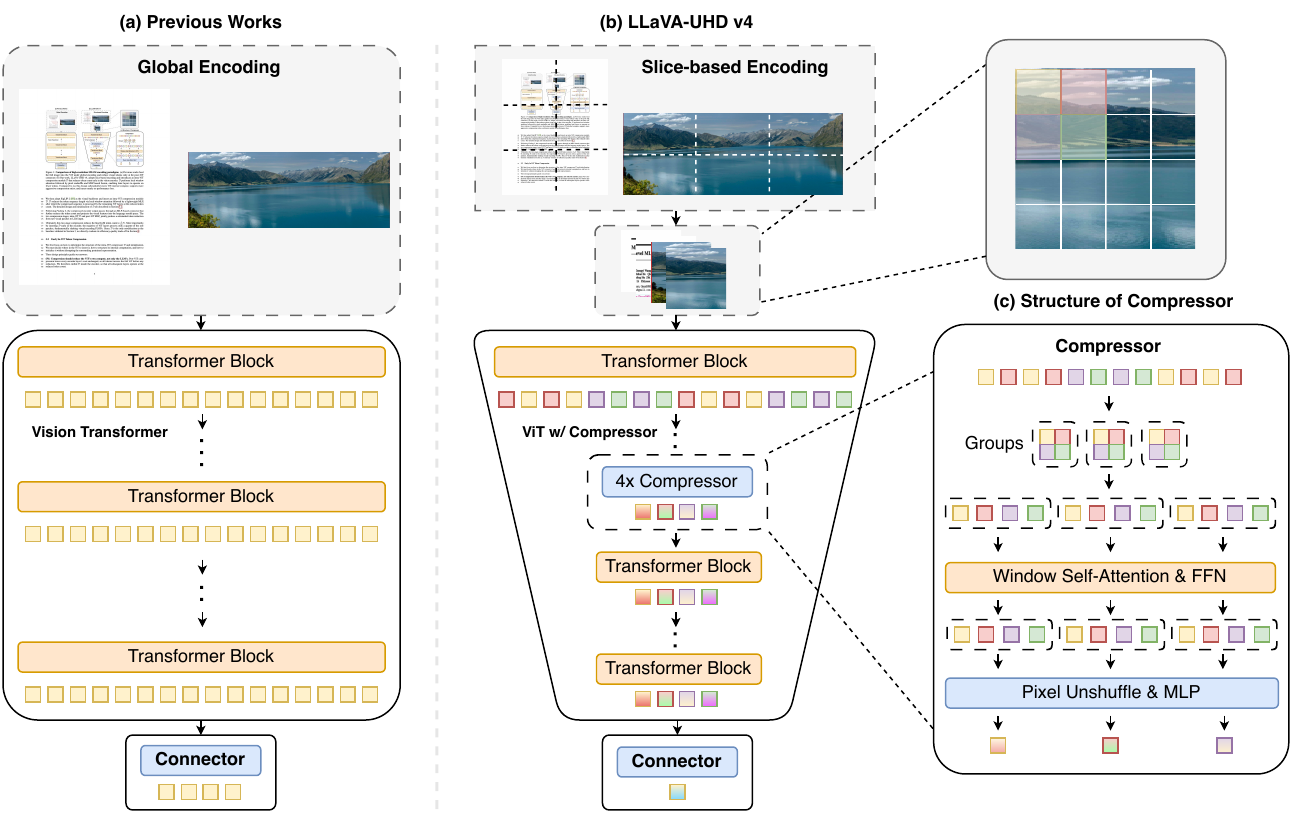}
    \caption{\textbf{Comparison of high-resolution MLLM encoding paradigms.} (a) Previous works feed the full image into the ViT under global encoding and reduce visual tokens only at the post-ViT connector. (b) Our work, LLaVA-UHD v4, adopts slice-based encoding and introduces an intra-ViT compression module $\mathcal{D}$ that reduces token count early in the vision encoder. $\mathcal{D}$ performs local window attention followed by pixel unshuffle and MLP-based fusion, enabling later layers to operate on fewer tokens. Compared to (a), this design substantially lowers ViT-internal compute, supports more aggressive compression ratios, and incurs nearly no performance loss.}
    \label{fig:architecture}
\end{figure}

Figure~\ref{fig:architecture} shows the full pipeline. Following \textit{Finding 1}, the input image is decomposed into a low-resolution thumbnail and a small set of high-resolution slices selected by an aspect-ratio-aware policy. All views are rescaled and concatenated along the sequence dimension, and processed in a single ViT forward pass that preserves per-view attention locality. 

We then adopt SigLIP 2~\cite{tschannen2025siglip} as the visual backbone and insert an intra-ViT compression module $\mathcal{D}$. $\mathcal{D}$ reduces the token sequence length via local window-attention followed by a lightweight MLP, after which the compressed sequence is processed by the remaining ViT layers at the reduced token count. The detailed design and initialization of $\mathcal{D}$ are described in Section~\ref{sec:early_compression}.

Following \textit{Finding 2}, the compressed encoder output passes through an MLP-based connector that further reduces the token count and projects the visual features into the language model space. The two compression stages, intra-ViT $\mathcal{D}$ and post-ViT MLP, jointly produce a substantial token reduction from raw visual patches to LLM input.

Ultimately, this two-stage compression reduces the final LLM token count to $\frac{1}{16}N$. More importantly, by inserting $\mathcal{D}$ early in the encoder, the majority of ViT layers process only a quarter of the raw patches, fundamentally slashing visual-encoding FLOPs. Since $\mathcal{D}$ is the only modification to the baseline validated in Section 2, we directly evaluate its efficiency-quality trade-off in Section~\ref{sec:experiment}.

\subsection{Early In-ViT Token Compression}
\label{sec:early_compression}

We first focus on determining the structure and initialization of the intra-ViT compressor $\mathcal{D}$. We must decide where in the ViT to insert it, how to structure its internal computation, and how to initialize it without disrupting the surrounding pretrained representation. 

Three design principles guide our answers. 

\textbf{(P1) Compression should reduce the ViT's own compute, not only the LLM's.} Post-ViT compression leaves every encoder layer's cost unchanged, as all tokens traverse the full ViT before any reduction. We therefore embed $\mathcal{D}$ inside the encoder, so that all subsequent layers operate at the reduced token count. 

\textbf{(P2) The compressor should sit as early as possible, balanced against representational depth.} Earlier insertion maximizes savings, while deeper placement retains more pretrained processing at full resolution and better aligns with the downstream representation manifold. Our ablations (Section~\ref{sec:ablation_compressor}) identify $k{=}6$ as the best efficiency-quality trade-off. 

\textbf{(P3) Inserting $\mathcal{D}$ must not disrupt the pretrained representation manifold.} A pretrained ViT is tightly calibrated, with each layer expecting the distribution produced by its predecessor. A randomly initialized $\mathcal{D}$ would perturb this distribution and turn fine-tuning into the harder problem of recovering the pretrained manifold from scratch. We therefore initialize $\mathcal{D}$ by reusing the parameters of the preceding ViT layer (Section~\ref{sec:initialization}), so that fine-tuning begins on the manifold rather than searching for it.

Together, these three principles fix $\mathcal{D}$'s placement and initialization strategy. It remains to specify the internal computation of $\mathcal{D}$ and the precise weight-inheritance mechanism, which we address in the rest of this section.

\subsubsection{Window-Attention Downsampling Module}
\label{sec:downsampling}

The pretrained ViT consists of $L$ transformer layers operating on token sequences $\mathbf{X}_l \in \mathbb{R}^{N \times d}$. We insert a downsampling module $\mathcal{D}$ between layers $k$ and $k{+}1$. The module takes $\mathbf{X}_k$ as input and produces a compressed sequence $\widetilde{\mathbf{X}} \in \mathbb{R}^{N/4 \times d}$, after which the remaining layers operate at the reduced token resolution. The module $\mathcal{D}$ consists of two conceptual steps: (i) a window-attention block that enriches local context, and (ii) a downsample-and-fuse block that reduces spatial resolution while aggregating information. 

\textbf{Window attention.} We first apply a window attention operator $\text{WinAttn}_{2\times2}$ on $\mathbf{X}_k$, producing an intermediate representation $\mathbf{Y}$. The attention is restricted to non-overlapping $2\times2$ windows, so each token interacts only with its three spatial neighbors. This design ensures that tokens exchange information exactly within the region that will be merged in the next step.

\textbf{Downsample and fuse.} A $2{\times}2$ PixelUnshuffle operation directly reshapes $\mathbf{Y}$ into $\mathbf{Z} \in \mathbb{R}^{N/4 \times 4d}$. An MLP then fuses these concatenated channels back to dimension $d$, yielding the final output $\widetilde{\mathbf{X}}$.

This design cleanly separates local context aggregation from information-preserving downsampling and channel fusion, while keeping the module compatible with the pretrained ViT stack.

\subsubsection{Parameter-Reuse Initialization}
\label{sec:initialization}

The downsampling module $\mathcal{D}$ introduces three parameterized components: the window-attention sub-block, the fused MLP $(\mathbf{W}_1, \phi, \mathbf{W}_2)$, and the two LayerNorms. A standard random initialization would inject substantial noise into the encoder's intermediate representations. In practice, this perturbation lengthens fine-tuning and is not guaranteed to recover the pretrained ViT's effective representation manifold at all.

We instead initialize $\mathcal{D}$ entirely from the weights of the pretrained ViT layer $k$ that immediately precedes it. This parameter reuse serves two purposes: it eliminates randomly-initialized parameters from the encoder's compute path entirely, and, as we make precise below, it places $\mathcal{D}$ at $t = 0$ in close functional correspondence to a surrogate operation derived from layer $k$ itself, so that fine-tuning starts on or near the pretrained representation manifold. We initialize $\mathcal{D}$ as follows:

\textbf{Window attention.} The attention projections, head configuration, and $\mathrm{LN}_1$ are copied directly from layer $k$. The only modification is the $2\times2$ window mask, which restricts attention to local neighborhoods while preserving the original attention weights.

\textbf{Fused MLP.} We construct the MLP to mimic applying the FFN of layer $k$ independently to each of the four patches within a $2\times2$ window, followed by averaging. Concretely,
\[
\mathbf{W}_1 = \mathrm{BlockDiag}(\mathbf{F}_1^{(k)}, \mathbf{F}_1^{(k)}, \mathbf{F}_1^{(k)}, \mathbf{F}_1^{(k)}), \quad
\mathbf{W}_2 = \tfrac{1}{4}[\mathbf{F}_2^{(k)} \mid \mathbf{F}_2^{(k)} \mid \mathbf{F}_2^{(k)} \mid \mathbf{F}_2^{(k)}].
\]
The bias follows the original FFN and is not scaled, so that the fused output corresponds to averaging four FFN branches while preserving the bias magnitude.

\textbf{LayerNorm and residual.} $\mathrm{LN}_2$ is applied over the concatenated $4d$ features with tiled affine parameters, and the residual branch is implemented as a parameter-free $2\times2$ average pooling.

\section{Experiment}
\label{sec:experiment}

\begin{figure}[t]
  \centering

  \begin{minipage}{0.48\linewidth}
    \centering
    \includegraphics[height=0.23\textheight]{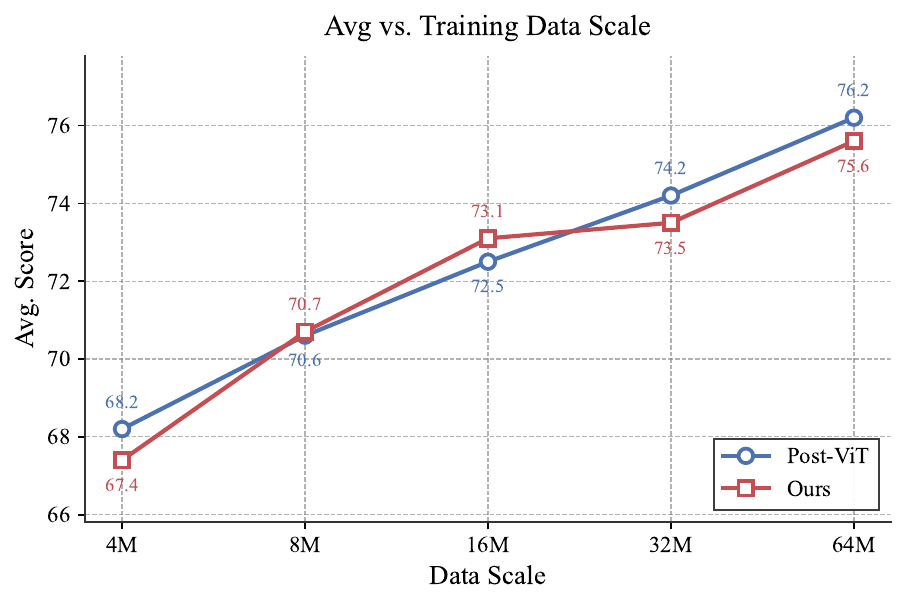}
  \end{minipage}
  \hfill
  \begin{minipage}{0.48\linewidth}
    \centering
    \includegraphics[height=0.23\textheight]{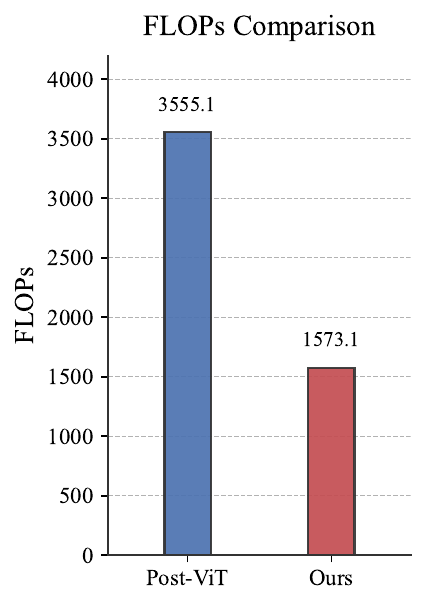}
  \end{minipage}

  \caption{\textbf{Average performance and computational cost.}
  Left: average accuracy across training data scales, comparing LLaVA-UHD v4 and the post-ViT baseline.
  Right: FLOPs comparison between the two systems.}
  \label{fig:avg_flops_wrap}
\end{figure}

We empirically validate the design of LLaVA-UHD v4 through controlled comparisons against the best-performing configuration from the pilot study (slice-based encoding with a $16\times$ post-ViT MLP 
compressor, hereafter the \emph{post-ViT baseline}). Section~\ref{sec:exp_setup} describes the setup, Section~\ref{sec:internal_closed} reports the main quality-efficiency results across training data scales, and Section~\ref{sec:ablation_compressor} analyzes the key design choices of the intra-ViT compressor.

\subsection{Experimental Setup}
\label{sec:exp_setup}

\textbf{Architecture.} Unless otherwise stated, LLaVA-UHD v4 uses SigLIP 2~\cite{tschannen2025siglip} as the vision encoder and Qwen3-8B~\cite{yang2025qwen3} as the language model. The intra-ViT compression module $\mathcal{D}$ is inserted after 
layer $k = 6$ and reduces the per-slice token count by $4\times$. A post-ViT MLP compressor further downsamples by $4\times$, yielding 
an end-to-end $16\times$ reduction. Unless otherwise stated, the FLOPs are computed for processing a single slice through the ViT, i.e., the visual-encoding cost per input slice.

\textbf{Training.} We follow a four-stage recipe: (i) \textit{Vision-language alignment} on large-scale image-text pairs, updating only the projector and $\mathcal{D}$; (ii) \textit{Knowledge injection} via OCR, document, and chart data with only ViT unfrozen; (iii) \textit{Interleaved training} on image-text sequences for multi-image and long-context reasoning; and (iv) \textit{Supervised instruction tuning} on a diverse mixture of general VQA, math, and conversational tasks. Detailed hyperparameters are in Appendix~\ref{app:hparams}.

\begin{figure}[t!]
  \centering
  \begin{subfigure}{0.24\linewidth}
    \centering
    \includegraphics[width=\linewidth]{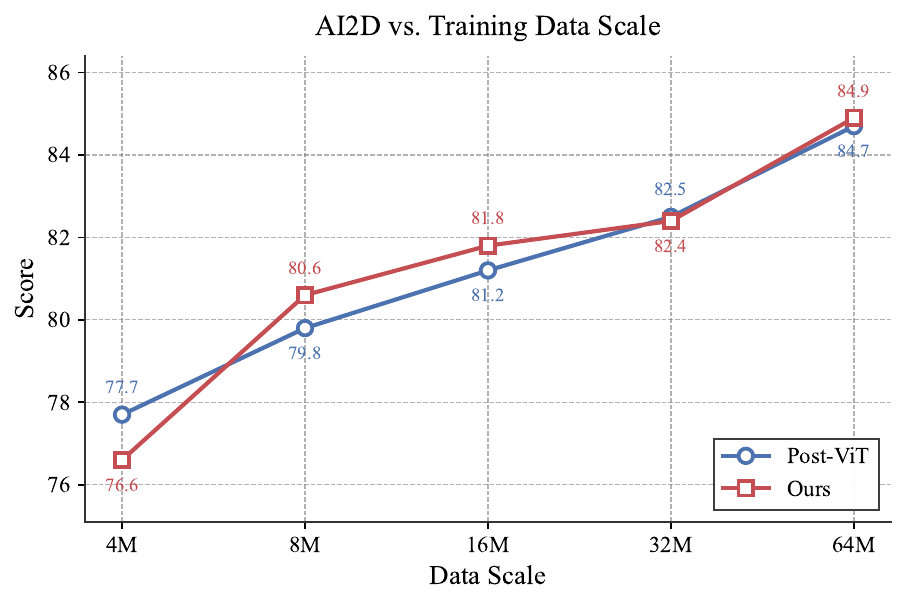}
    \caption{AI2D}
  \end{subfigure}
  \begin{subfigure}{0.24\linewidth}
    \centering
    \includegraphics[width=\linewidth]{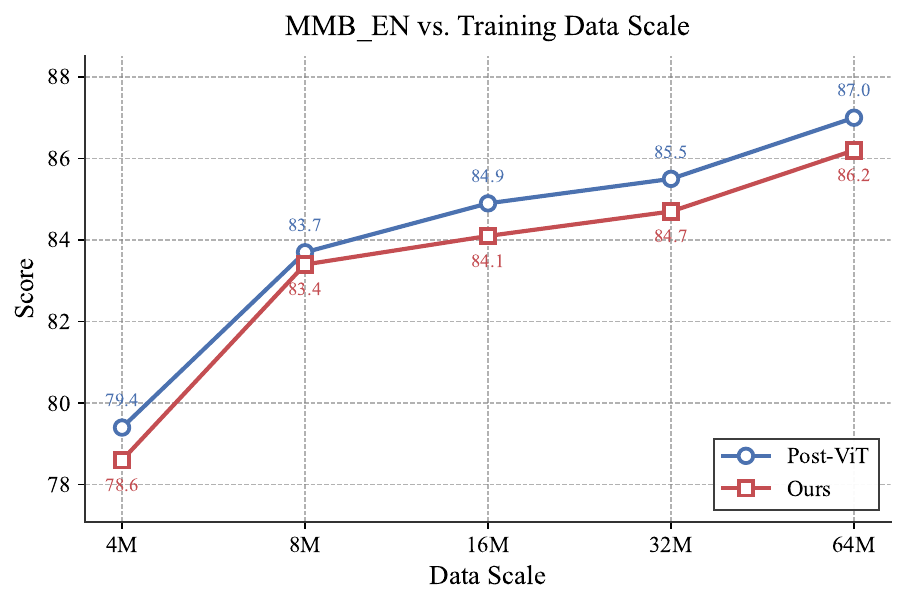}
    \caption{MMBench$_\text{EN}$}
  \end{subfigure}
  \begin{subfigure}{0.24\linewidth}
    \centering
    \includegraphics[width=\linewidth]{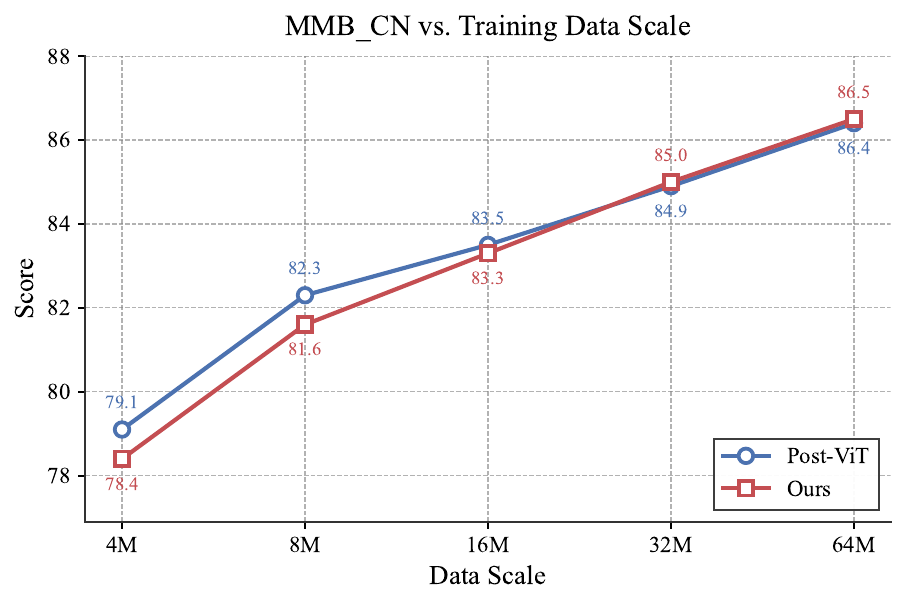}
    \caption{MMBench$_\text{CN}$}
  \end{subfigure}
  \begin{subfigure}{0.24\linewidth}
    \centering
    \includegraphics[width=\linewidth]{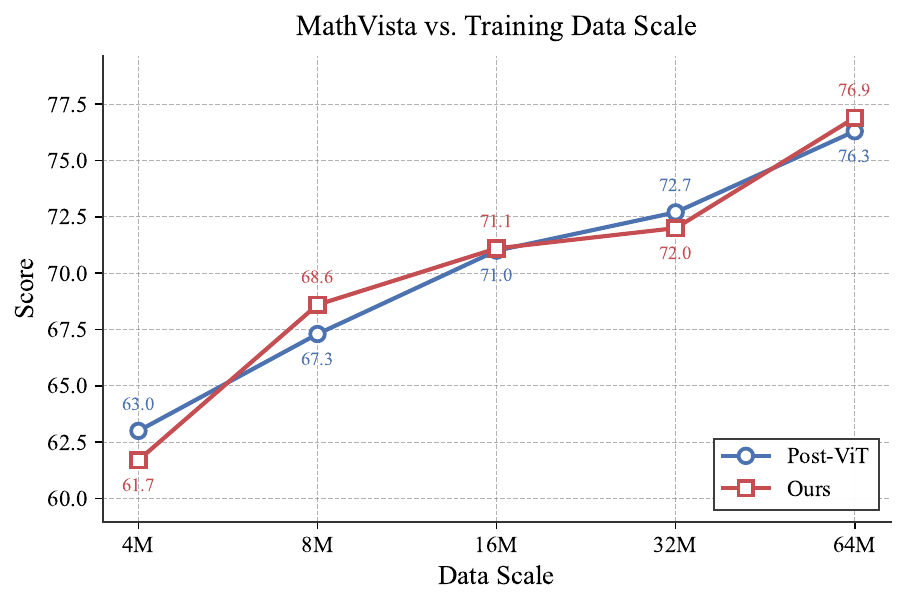}
    \caption{MathVista}
  \end{subfigure}

  \vspace{0.5em}

  \begin{subfigure}{0.24\linewidth}
    \centering
    \includegraphics[width=\linewidth]{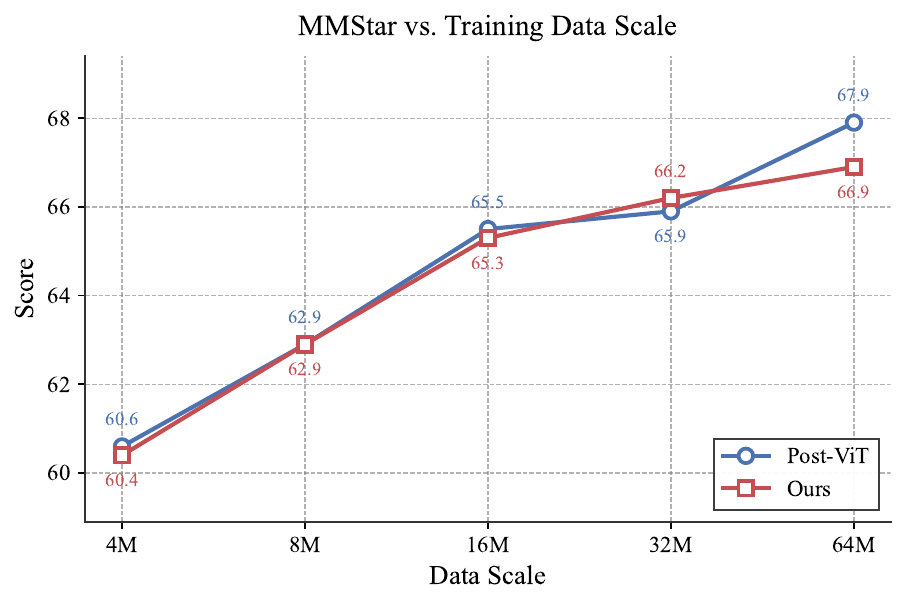}
    \caption{MMStar}
  \end{subfigure}
  \begin{subfigure}{0.24\linewidth}
    \centering
    \includegraphics[width=\linewidth]{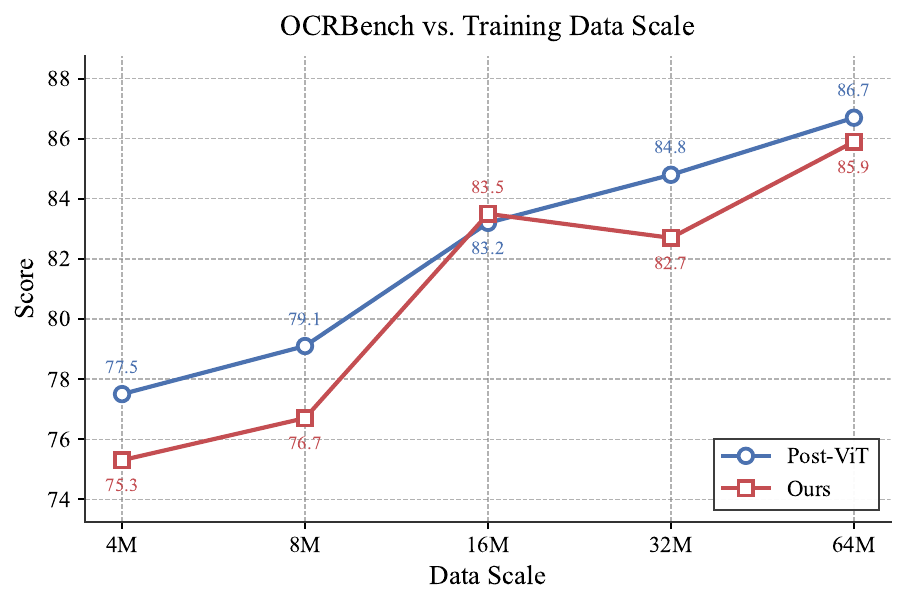}
    \caption{OCRBench}
  \end{subfigure}
  \begin{subfigure}{0.24\linewidth}
    \centering
    \includegraphics[width=\linewidth]{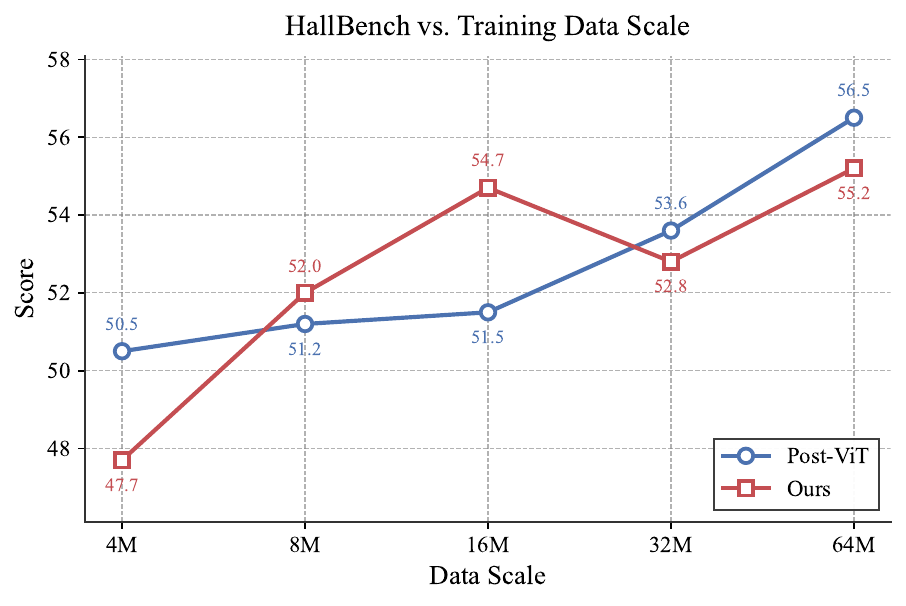}
    \caption{HallBench}
  \end{subfigure}
  \begin{subfigure}{0.24\linewidth}
    \centering
    \includegraphics[width=\linewidth]{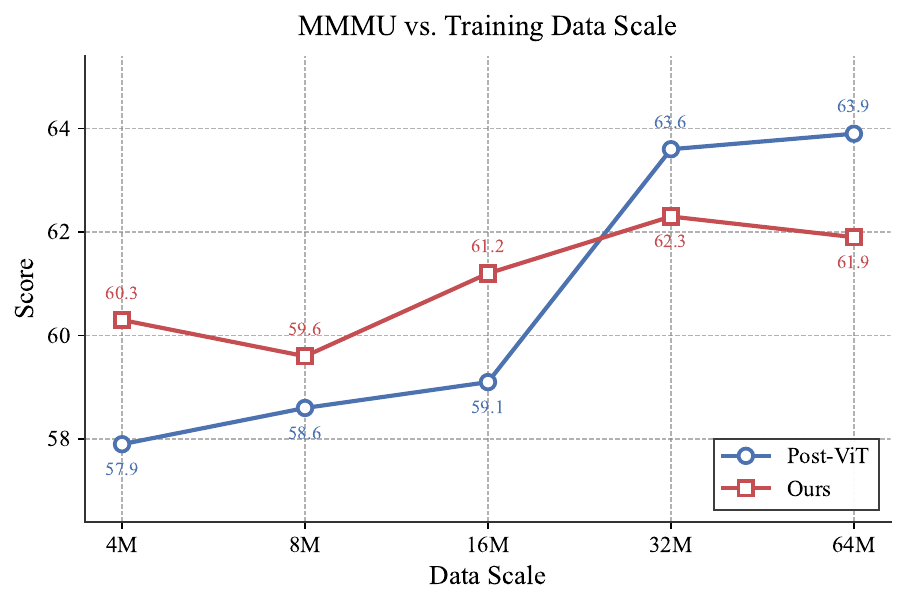}
    \caption{MMMU}
  \end{subfigure}

  \caption{\textbf{Benchmark trends across training data scales.}
  We compare Post-ViT and our method on eight benchmarks across different training data scales.}
  \label{fig:benchmark_trends}
\end{figure}

\textbf{Benchmarks.} We evaluate on eight benchmarks covering three capability dimensions: (i) \emph{general VQA}: MMBench$_\text{EN}$~\cite{liu2024mmbench}, MMBench$_\text{CN}$~\cite{liu2024mmbench}, MMStar~\cite{chen2024we}; (ii) \emph{knowledge \& reasoning}: MMMU~\cite{yue2024mmmu}, MathVista~\cite{lu2023mathvista}, AI2D~\cite{kembhavi2016diagram}, HallusionBench~\cite{guan2024hallusionbench}; (iii) \emph{fine-grained perception}: OCRBench~\cite{liu2024ocrbench}. 

\subsection{Main Results}
\label{sec:internal_closed}

\textbf{Intra-ViT early compression matches the post-ViT baseline in accuracy while substantially reducing visual-encoding cost.}
As shown in Figure~\ref{fig:avg_flops_wrap} and Figure~\ref{fig:benchmark_trends}, we compare LLaVA-UHD v4 against the strongest post-ViT baseline under identical training settings and a shared end-to-end $16\times$ compression ratio. By shifting a $4\times$ compression stage inside the ViT, all subsequent layers operate on only $25\%$ of the original tokens. This structurally reduces visual-encoding FLOPs from $3555$G to $1573$G, a massive $55.75\%$ reduction. Despite this aggressive early compression, LLaVA-UHD v4 performs within $\pm 0.8$ points of the baseline across all five training scales, with a negligible mean deviation of only $-0.29$ points. This demonstrates that our intra-ViT design yields massive compute savings without compromising downstream accuracy.

\textbf{The proposed early-compression design preserves average scaling behavior within the tested range.}
As training data increases from 4M to 64M samples, both systems improve substantially. The post-ViT baseline rises from $68.2$ to $76.2$ average points, while LLaVA-UHD v4 rises from $67.4$ to $75.6$. The average gap stays within $\pm 0.8$ points and does not widen monotonically, suggesting that intra-ViT compression does not introduce an observable average-level scaling ceiling. Individual benchmarks still show scale-dependent variation, for example, MMMU favors LLaVA-UHD v4 at smaller scales but the post-ViT baseline at larger scales, but this reversal does not indicate a systematic compression failure, since the aggregate trend remains stable across the tested range.

\subsection{Ablations on the In-ViT Compression Module}
\label{sec:ablation_compressor}

Section~\ref{sec:internal_closed} shows that LLaVA-UHD v4 can match the post-ViT baseline under the same final token budget. We now ablate the design of the intra-ViT compression module $\mathcal{D}$ to understand why this is possible. All variants use the 8M in-house training set and an end-to-end $16\times$ compression ratio, with $\mathcal{D}$ inserted at $k=6$ by default, applying $4\times$ reduction over $2\times 2$ token windows. \emph{Average Pool} and \emph{Pixel-Unshuffle} are parameter-free or randomly initialized merging baselines. \emph{Cross-Attn} collapses each window into one token via cross-attention with either the top-left or mean query.\emph{Win-Attn} variants first apply window self-attention and then fuse tokens with a Pixel-Unshuffle MLP, either randomly initialized or reused from the preceding ViT FFN. The central question is therefore not whether early compression can reduce compute, but which compressor can preserve the pretrained ViT representation while doing so.

\begin{table}[htbp]
\centering
\caption{\textbf{Ablations on in-ViT compression designs.}
All variants use the same final $16\times$ compression ratio and insertion depth $k=6$.}
\label{tab:ablation_grouped}
\setlength{\tabcolsep}{2.5pt}
\scriptsize

\begin{subtable}{0.3\linewidth}
\centering
\caption{Naive merging}
\begin{tabular}{l|cc}
\toprule
Method & FLOPs (G) & Avg. \\
\midrule
Post-ViT Base & 3555.1 & \textbf{70.6} \\
Avg Pool & \textbf{1368.7} & 69.6 \\
Pix-Unshuffle & 1401.2 & 69.8 \\
\bottomrule
\end{tabular}
\end{subtable}
\hfill
\begin{subtable}{0.32\linewidth}
\centering
\caption{Direct cross-attention}
\begin{tabular}{l|cc}
\toprule
Method & FLOPs (G) & Avg. \\
\midrule
Post-ViT Base & 3555.1 & \textbf{70.6} \\
Cross (top-left) & \textbf{1402.0} & 70.5 \\
Cross (mean) & \textbf{1402.0} & 69.9 \\
\bottomrule
\end{tabular}
\end{subtable}
\hfill
\begin{subtable}{0.35\linewidth}
\centering
\caption{Reused MLP and window attention}
\begin{tabular}{l|cc}
\toprule
Method & FLOPs (G) & Avg. \\
\midrule
Pix-Unshuffle & \textbf{1401.2} & 69.8 \\
Reused MLP & 1490.2 & 69.9 \\
Win w/ MLP & 1484.1 & 70.1 \\
Win w/ Reused& 1573.1 & \textbf{70.7} \\
\bottomrule
\end{tabular}
\end{subtable}

\end{table}

\textbf{Naive in-ViT compression is efficient but not sufficient.}
Table~\ref{tab:ablation_grouped}(a) first evaluates simple in-ViT merging strategies. Moving compression into the ViT substantially reduces computation, from $3555.1$G FLOPs for the post-ViT baseline to $1401.2$G FLOPs for in-ViT variants. However, this efficiency gain does not automatically recover baseline-level accuracy. Average pooling is the cheapest design, but drops the average score from $70.6$ to $69.6$. A learnable pixel-unshuffle MLP improves the score to $69.8$, but still remains below the post-ViT baseline. These results suggest that early token reduction creates a nontrivial interface problem within the pretrained ViT, requiring the compressor to reduce sequence length while maintaining compatibility with the representational distribution expected by the remaining encoder layers.

\textbf{Window attention and reuse initialization are complementary components of the structured merger.}
Table~\ref{tab:ablation_grouped}(c) factorizes our structured merger along two axes, whether local window attention is applied before merging, and whether the fusion MLP is initialized by reusing the preceding ViT FFN weights. Reuse alone brings only a marginal gain over a randomly initialized pixel-unshuffle MLP, improving the average score from $69.8$ to $69.9$. Window attention alone is more helpful, raising the score to $70.1$. When the two are combined, the score reaches $70.7$, exceeding both individual modifications and slightly surpassing the post-ViT baseline. The gain is super-additive because the two components together make the merger closely resemble a standard vision encoder block, with local self-attention followed by an FFN and both initialized from the preceding layer's weights. The output of the merger therefore stays close to what the subsequent ViT layers were pretrained to consume. Neither component alone achieves this alignment. Without window attention, the reused MLP is applied to tokens that have not been locally contextualized as in pretraining, so its initialization provides little benefit. Without reuse, window attention restores local structure but the randomly initialized fusion then maps the contextualized tokens out of the pretrained input distribution.

\begin{wraptable}{r}{0.4\linewidth}
  \vspace{-8pt}
  \centering
  \small
  \caption{\textbf{Effect of insertion depth $k$ on accuracy and compute.} Evaluation for $\mathcal{D}$ inserted after different ViT layers, reporting average score and visual-encoding FLOPs.}
  \label{tab:insertion_depth}
  \resizebox{\linewidth}{!}{
  \begin{tabular}{c|cc}
  \toprule
  Layer ($k$) & FLOPs (G) & Avg. Score  \\
  \midrule
  3  & \textbf{1245.1} & 39.7  \\
  6  & 1573.1 & \textbf{70.7} \\
  9  & 1901.1 & 70.3  \\
  15 & 2557.0 & 70.4 \\
  \bottomrule
  \end{tabular}}
  \vspace{-6pt}
\end{wraptable}

\textbf{Direct cross-attention merging underperforms local window attention followed by a reuse-initialized MLP.}
Table~\ref{tab:ablation_grouped}(b) compares against a more direct alternative that collapses each $2\times 2$ window into a single token through local cross-attention. This alternative is competitive when the top-left token is used as the query, reaching $70.5$ average accuracy, close to both the post-ViT baseline and our final design. However, changing the query to the window mean lowers the score to $69.8$ under the same FLOPs, showing that direct one-step aggregation is sensitive to how the representative query is constructed. In contrast, first updating all tokens through local window attention and then fusing the contextualized tokens with a reuse-initialized MLP achieves $70.7$, the best among all ablated in-ViT compressors. As shown in Table~\ref{tab:ablation_scale_inhouse}, this query sensitivity persists at $16$M, where the better-performing query even flips to the window mean, while Win-Attn with Reused MLP stays strongest at both scales. This suggests a structural issue rather than a tuning artifact, since no single query consistently captures what a $2\times 2$ window should be summarized into, whereas updating all tokens before fusion sidesteps the question entirely.

\textbf{Effective intra-ViT compression requires an intermediate insertion depth.}
As shown in Table~\ref{tab:insertion_depth}, inserting $\mathcal{D}$ too early is highly destructive: $k=3$ gives the lowest FLOPs, but drops the average score to $38.76$. This indicates that the earliest ViT layers have not yet formed representations that are safe to merge. In contrast, inserting at $k=6$ preserves accuracy while retaining most of the compute savings. Delaying compression to $k=9$ or $k=15$ brings no accuracy benefit, yielding slightly lower scores while increasing FLOPs to $1901$G and $2557$G, respectively. Among the non-collapsed settings in our sweep, $k=6$ is therefore Pareto-favorable. It is both more accurate and more efficient than the deeper insertion depths. This suggests that effective intra-ViT compression requires an intermediate depth where tokens are no longer purely low-level visual features but have already accumulated enough semantic structure to be safely merged.

\section{Conclusion}

In this work, we present LLaVA-UHD v4, a highly efficient visual encoding architecture that systematically re-examines high-resolution perception in MLLMs. By demonstrating the empirical advantages of slice-based encoding over the global encoding paradigm, and introducing a novel parameter-reusing intra-ViT early compression module, we substantially reduce the severe computational bottleneck inside the vision encoder. Extensive experiments validate that our approach reduces visual-encoding FLOPs by $55.75\%$ under a $16\times$ compression ratio, while matching or surpassing the fine-grained downstream performance of strong post-ViT baselines. While our current module operates at a fixed compression rate, exploring dynamic, content-aware token reduction mechanisms within the encoder remains an exciting direction for future research. Together, these results suggest that aggressive token reduction can be performed inside the vision encoder without sacrificing fine-grained perception, offering a practical path toward more scalable multimodal foundation models.

{
\small
  \bibliographystyle{plain}
  \bibliography{references}

@article{liu2023visual,
  title={Visual instruction tuning},
  author={Liu, Haotian and Li, Chunyuan and Wu, Qingyang and Lee, Yong Jae},
  journal={Advances in neural information processing systems},
  volume={36},
  pages={34892--34916},
  year={2023}
}

@inproceedings{liu2024improved,
  title={Improved baselines with visual instruction tuning},
  author={Liu, Haotian and Li, Chunyuan and Li, Yuheng and Lee, Yong Jae},
  booktitle={Proceedings of the IEEE/CVF conference on computer vision and pattern recognition},
  pages={26296--26306},
  year={2024}
}

@misc{bai2023qwenvlversatilevisionlanguagemodel,
      title={Qwen-VL: A Versatile Vision-Language Model for Understanding, Localization, Text Reading, and Beyond}, 
      author={Jinze Bai and Shuai Bai and Shusheng Yang and Shijie Wang and Sinan Tan and Peng Wang and Junyang Lin and Chang Zhou and Jingren Zhou},
      year={2023},
      eprint={2308.12966},
      archivePrefix={arXiv},
      primaryClass={cs.CV},
      url={https://arxiv.org/abs/2308.12966}, 
}

@article{yao2024minicpm,
  title={Minicpm-v: A gpt-4v level mllm on your phone},
  author={Yao, Yuan and Yu, Tianyu and Zhang, Ao and Wang, Chongyi and Cui, Junbo and Zhu, Hongji and Cai, Tianchi and Li, Haoyu and Zhao, Weilin and He, Zhihui and others},
  journal={arXiv preprint arXiv:2408.01800},
  year={2024}
}

@inproceedings{cha2024honeybee,
  title={Honeybee: Locality-enhanced projector for multimodal llm},
  author={Cha, Junbum and Kang, Wooyoung and Mun, Jonghwan and Roh, Byungseok},
  booktitle={Proceedings of the IEEE/CVF Conference on Computer Vision and Pattern Recognition},
  pages={13817--13827},
  year={2024}
}

@inproceedings{guo2024llava,
  title={Llava-uhd: an lmm perceiving any aspect ratio and high-resolution images},
  author={Guo, Zonghao and Xu, Ruyi and Yao, Yuan and Cui, Junbo and Ni, Zanlin and Ge, Chunjiang and Chua, Tat-Seng and Liu, Zhiyuan and Huang, Gao},
  booktitle={European Conference on Computer Vision},
  pages={390--406},
  year={2024},
  organization={Springer}
}

@inproceedings{chen2024internvl,
  title={Internvl: Scaling up vision foundation models and aligning for generic visual-linguistic tasks},
  author={Chen, Zhe and Wu, Jiannan and Wang, Wenhai and Su, Weijie and Chen, Guo and Xing, Sen and Zhong, Muyan and Zhang, Qinglong and Zhu, Xizhou and Lu, Lewei and others},
  booktitle={Proceedings of the IEEE/CVF conference on computer vision and pattern recognition},
  pages={24185--24198},
  year={2024}
}

@misc{liu2024llavanext,
    title={LLaVA-NeXT: Improved reasoning, OCR, and world knowledge},
    url={https://llava-vl.github.io/blog/2024-01-30-llava-next/},
    author={Liu, Haotian and Li, Chunyuan and Li, Yuheng and Li, Bo and Zhang, Yuanhan and Shen, Sheng and Lee, Yong Jae},
    month={January},
    year={2024}
}

@article{wang2024qwen2,
  title={Qwen2-vl: Enhancing vision-language model's perception of the world at any resolution},
  author={Wang, Peng and Bai, Shuai and Tan, Sinan and Wang, Shijie and Fan, Zhihao and Bai, Jinze and Chen, Keqin and Liu, Xuejing and Wang, Jialin and Ge, Wenbin and others},
  journal={arXiv preprint arXiv:2409.12191},
  year={2024}
}

@article{lu2025internvl,
  title={Internvl-x: Advancing and accelerating internvl series with efficient visual token compression},
  author={Lu, Dongchen and Sun, Yuyao and Zhang, Zilu and Huang, Leping and Zeng, Jianliang and Shu, Mao and Cao, Huo},
  journal={arXiv preprint arXiv:2503.21307},
  year={2025}
}

@inproceedings{li2023blip,
  title={Blip-2: Bootstrapping language-image pre-training with frozen image encoders and large language models},
  author={Li, Junnan and Li, Dongxu and Savarese, Silvio and Hoi, Steven},
  booktitle={International conference on machine learning},
  pages={19730--19742},
  year={2023},
  organization={PMLR}
}

@inproceedings{Mathew_2021_WACV,
    author = {Mathew, Minesh and Karatzas, Dimosthenis and Jawahar, C.V.},
    booktitle = {Proceedings of the IEEE/CVF Winter Conference on Applications of Computer Vision (WACV)},
    pages = {2200-2209},
    title = {{DocVQA: A Dataset for VQA on Document Images}},
    year = {2021}
}

@inproceedings{ouyang2025omnidocbench,
  title={Omnidocbench: Benchmarking diverse pdf document parsing with comprehensive annotations},
  author={Ouyang, Linke and Qu, Yuan and Zhou, Hongbin and Zhu, Jiawei and Zhang, Rui and Lin, Qunshu and Wang, Bin and Zhao, Zhiyuan and Jiang, Man and Zhao, Xiaomeng and others},
  booktitle={Proceedings of the IEEE/CVF Conference on Computer Vision and Pattern Recognition},
  pages={24838--24848},
  year={2025}
}

@inproceedings{masry2022chartqa,
  title={Chartqa: A benchmark for question answering about charts with visual and logical reasoning},
  author={Masry, Ahmed and Do, Xuan Long and Tan, Jia Qing and Joty, Shafiq and Hoque, Enamul},
  booktitle={Findings of the association for computational linguistics: ACL 2022},
  pages={2263--2279},
  year={2022}
}

@article{zhang2024mme,
  title={Mme-realworld: Could your multimodal llm challenge high-resolution real-world scenarios that are difficult for humans?},
  author={Zhang, Yi-Fan and Zhang, Huanyu and Tian, Haochen and Fu, Chaoyou and Zhang, Shuangqing and Wu, Junfei and Li, Feng and Wang, Kun and Wen, Qingsong and Zhang, Zhang and others},
  journal={arXiv preprint arXiv:2408.13257},
  year={2024}
}

@article{dai2023instructblip,
  title={Instructblip: Towards general-purpose vision-language models with instruction tuning},
  author={Dai, Wenliang and Li, Junnan and Li, Dongxu and Tiong, Anthony and Zhao, Junqi and Wang, Weisheng and Li, Boyang and Fung, Pascale N and Hoi, Steven},
  journal={Advances in neural information processing systems},
  volume={36},
  pages={49250--49267},
  year={2023}
}

@inproceedings{wu2024v,
  title={V?: Guided visual search as a core mechanism in multimodal llms},
  author={Wu, Penghao and Xie, Saining},
  booktitle={Proceedings of the IEEE/CVF Conference on Computer Vision and Pattern Recognition},
  pages={13084--13094},
  year={2024}
}

@inproceedings{radford2021learning,
  title={Learning transferable visual models from natural language supervision},
  author={Radford, Alec and Kim, Jong Wook and Hallacy, Chris and Ramesh, Aditya and Goh, Gabriel and Agarwal, Sandhini and Sastry, Girish and Askell, Amanda and Mishkin, Pamela and Clark, Jack and others},
  booktitle={International conference on machine learning},
  pages={8748--8763},
  year={2021},
  organization={PmLR}
}

@inproceedings{zhai2023sigmoid,
  title={Sigmoid loss for language image pre-training},
  author={Zhai, Xiaohua and Mustafa, Basil and Kolesnikov, Alexander and Beyer, Lucas},
  booktitle={Proceedings of the IEEE/CVF international conference on computer vision},
  pages={11975--11986},
  year={2023}
}

@article{ilharco2021openclip,
  title={Openclip},
  author={Ilharco, Gabriel and Wortsman, Mitchell and Carlini, Nicholas and Taori, Rohan and Dave, Achal and Shankar, Vaishaal and Namkoong, Hongseok and Miller, John and Hajishirzi, Hannaneh and Farhadi, Ali and others},
  journal={Zenodo},
  year={2021}
}

@article{xu2023demystifying,
  title={Demystifying clip data},
  author={Xu, Hu and Xie, Saining and Tan, Xiaoqing Ellen and Huang, Po-Yao and Howes, Russell and Sharma, Vasu and Li, Shang-Wen and Ghosh, Gargi and Zettlemoyer, Luke and Feichtenhofer, Christoph},
  journal={arXiv preprint arXiv:2309.16671},
  year={2023}
}

@article{sun2023eva,
  title={Eva-clip: Improved training techniques for clip at scale},
  author={Sun, Quan and Fang, Yuxin and Wu, Ledell and Wang, Xinlong and Cao, Yue},
  journal={arXiv preprint arXiv:2303.15389},
  year={2023}
}

@article{tschannen2025siglip,
  title={Siglip 2: Multilingual vision-language encoders with improved semantic understanding, localization, and dense features},
  author={Tschannen, Michael and Gritsenko, Alexey and Wang, Xiao and Naeem, Muhammad Ferjad and Alabdulmohsin, Ibrahim and Parthasarathy, Nikhil and Evans, Talfan and Beyer, Lucas and Xia, Ye and Mustafa, Basil and others},
  journal={arXiv preprint arXiv:2502.14786},
  year={2025}
}

@inproceedings{fini2025multimodal,
  title={Multimodal autoregressive pre-training of large vision encoders},
  author={Fini, Enrico and Shukor, Mustafa and Li, Xiujun and Dufter, Philipp and Klein, Michal and Haldimann, David and Aitharaju, Sai and da Costa, Victor G Turrisi and B{\'e}thune, Louis and Gan, Zhe and others},
  booktitle={Proceedings of the IEEE/CVF Conference on Computer Vision and Pattern Recognition},
  pages={9641--9654},
  year={2025}
}

@article{dehghani2023patch,
  title={Patch n’pack: Navit, a vision transformer for any aspect ratio and resolution},
  author={Dehghani, Mostafa and Mustafa, Basil and Djolonga, Josip and Heek, Jonathan and Minderer, Matthias and Caron, Mathilde and Steiner, Andreas and Puigcerver, Joan and Geirhos, Robert and Alabdulmohsin, Ibrahim M and others},
  journal={Advances in Neural Information Processing Systems},
  volume={36},
  pages={2252--2274},
  year={2023}
}

@article{alayrac2022flamingo,
  title={Flamingo: a visual language model for few-shot learning},
  author={Alayrac, Jean-Baptiste and Donahue, Jeff and Luc, Pauline and Miech, Antoine and Barr, Iain and Hasson, Yana and Lenc, Karel and Mensch, Arthur and Millican, Katherine and Reynolds, Malcolm and others},
  journal={Advances in neural information processing systems},
  volume={35},
  pages={23716--23736},
  year={2022}
}

@article{zhu2023minigpt,
  title={Minigpt-4: Enhancing vision-language understanding with advanced large language models},
  author={Zhu, Deyao and Chen, Jun and Shen, Xiaoqian and Li, Xiang and Elhoseiny, Mohamed},
  journal={arXiv preprint arXiv:2304.10592},
  year={2023}
}

@article{huang2023language,
  title={Language is not all you need: Aligning perception with language models},
  author={Huang, Shaohan and Dong, Li and Wang, Wenhui and Hao, Yaru and Singhal, Saksham and Ma, Shuming and Lv, Tengchao and Cui, Lei and Mohammed, Owais Khan and Patra, Barun and others},
  journal={Advances in Neural Information Processing Systems},
  volume={36},
  pages={72096--72109},
  year={2023}
}

@article{peng2023kosmos,
  title={Kosmos-2: Grounding multimodal large language models to the world},
  author={Peng, Zhiliang and Wang, Wenhui and Dong, Li and Hao, Yaru and Huang, Shaohan and Ma, Shuming and Wei, Furu},
  journal={arXiv preprint arXiv:2306.14824},
  year={2023}
}

@article{ye2023mplug,
  title={mplug-owl: Modularization empowers large language models with multimodality},
  author={Ye, Qinghao and Xu, Haiyang and Xu, Guohai and Ye, Jiabo and Yan, Ming and Zhou, Yiyang and Wang, Junyang and Hu, Anwen and Shi, Pengcheng and Shi, Yaya and others},
  journal={arXiv preprint arXiv:2304.14178},
  year={2023}
}

@inproceedings{chen2024sharegpt4v,
  title={Sharegpt4v: Improving large multi-modal models with better captions},
  author={Chen, Lin and Li, Jinsong and Dong, Xiaoyi and Zhang, Pan and He, Conghui and Wang, Jiaqi and Zhao, Feng and Lin, Dahua},
  booktitle={European Conference on Computer Vision},
  pages={370--387},
  year={2024},
  organization={Springer}
}

@article{li2025mini,
  title={Mini-gemini: Mining the potential of multi-modality vision language models},
  author={Li, Yanwei and Zhang, Yuechen and Wang, Chengyao and Zhong, Zhisheng and Chen, Yixin and Chu, Ruihang and Liu, Shaoteng and Jia, Jiaya},
  journal={IEEE Transactions on Pattern Analysis and Machine Intelligence},
  year={2025},
  publisher={IEEE}
}

@inproceedings{karamcheti2024prismatic,
  title     = {Prismatic {VLMs}: Investigating the Design Space of Visually-Conditioned Language Models},
  author    = {Karamcheti, Siddharth and Nair, Suraj and Balakrishna, Ashwin and Liang, Percy and Kollar, Thomas and Sadigh, Dorsa},
  booktitle = {International Conference on Machine Learning (ICML)},
  year      = {2024}
}

@article{wang2024cogvlm,
  title={Cogvlm: Visual expert for pretrained language models},
  author={Wang, Weihan and Lv, Qingsong and Yu, Wenmeng and Hong, Wenyi and Qi, Ji and Wang, Yan and Ji, Junhui and Yang, Zhuoyi and Zhao, Lei and Song, Xixuan and others},
  journal={Advances in Neural Information Processing Systems},
  volume={37},
  pages={121475--121499},
  year={2024}
}

@inproceedings{mckinzie2024mm1,
  title={Mm1: methods, analysis and insights from multimodal llm pre-training},
  author={McKinzie, Brandon and Gan, Zhe and Fauconnier, Jean-Philippe and Dodge, Sam and Zhang, Bowen and Dufter, Philipp and Shah, Dhruti and Du, Xianzhi and Peng, Futang and Belyi, Anton and others},
  booktitle={European Conference on Computer Vision},
  pages={304--323},
  year={2024},
  organization={Springer}
}

@article{team2026kimi,
  title={Kimi K2. 5: Visual Agentic Intelligence},
  author={Team, Kimi and Bai, Tongtong and Bai, Yifan and Bao, Yiping and Cai, SH and Cao, Yuan and Charles, Y and Che, HS and Chen, Cheng and Chen, Guanduo and others},
  journal={arXiv preprint arXiv:2602.02276},
  year={2026}
}

@article{bolya2022token,
  title={Token merging: Your vit but faster},
  author={Bolya, Daniel and Fu, Cheng-Yang and Dai, Xiaoliang and Zhang, Peizhao and Feichtenhofer, Christoph and Hoffman, Judy},
  journal={arXiv preprint arXiv:2210.09461},
  year={2022}
}

@article{chen2024far,
  title={How far are we to gpt-4v? closing the gap to commercial multimodal models with open-source suites},
  author={Chen, Zhe and Wang, Weiyun and Tian, Hao and Ye, Shenglong and Gao, Zhangwei and Cui, Erfei and Tong, Wenwen and Hu, Kongzhi and Luo, Jiapeng and Ma, Zheng and others},
  journal={Science China Information Sciences},
  volume={67},
  number={12},
  pages={220101},
  year={2024},
  publisher={Springer}
}

@inproceedings{hu2024mplug,
  title={mplug-docowl 1.5: Unified structure learning for ocr-free document understanding},
  author={Hu, Anwen and Xu, Haiyang and Ye, Jiabo and Yan, Ming and Zhang, Liang and Zhang, Bo and Zhang, Ji and Jin, Qin and Huang, Fei and Zhou, Jingren},
  booktitle={Findings of the Association for Computational Linguistics: EMNLP 2024},
  pages={3096--3120},
  year={2024}
}

@inproceedings{chen2024image,
  title={An image is worth 1/2 tokens after layer 2: Plug-and-play inference acceleration for large vision-language models},
  author={Chen, Liang and Zhao, Haozhe and Liu, Tianyu and Bai, Shuai and Lin, Junyang and Zhou, Chang and Chang, Baobao},
  booktitle={European Conference on Computer Vision},
  pages={19--35},
  year={2024},
  organization={Springer}
}

@article{zhang2024sparsevlm,
  title={Sparsevlm: Visual token sparsification for efficient vision-language model inference},
  author={Zhang, Yuan and Fan, Chun-Kai and Ma, Junpeng and Zheng, Wenzhao and Huang, Tao and Cheng, Kuan and Gudovskiy, Denis and Okuno, Tomoyuki and Nakata, Yohei and Keutzer, Kurt and others},
  journal={arXiv preprint arXiv:2410.04417},
  year={2024}
}

@inproceedings{lin2025boosting,
  title={Boosting multimodal large language models with visual tokens withdrawal for rapid inference},
  author={Lin, Zhihang and Lin, Mingbao and Lin, Luxi and Ji, Rongrong},
  booktitle={Proceedings of the AAAI Conference on Artificial Intelligence},
  volume={39},
  number={5},
  pages={5334--5342},
  year={2025}
}

@article{xing2024pyramiddrop,
  title={Pyramiddrop: Accelerating your large vision-language models via pyramid visual redundancy reduction},
  author={Xing, Long and Huang, Qidong and Dong, Xiaoyi and Lu, Jiajie and Zhang, Pan and Zang, Yuhang and Cao, Yuhang and He, Conghui and Wang, Jiaqi and Wu, Feng and others},
  journal={arXiv preprint arXiv:2410.17247},
  year={2024}
}

@article{rao2021dynamicvit,
  title={Dynamicvit: Efficient vision transformers with dynamic token sparsification},
  author={Rao, Yongming and Zhao, Wenliang and Liu, Benlin and Lu, Jiwen and Zhou, Jie and Hsieh, Cho-Jui},
  journal={Advances in neural information processing systems},
  volume={34},
  pages={13937--13949},
  year={2021}
}

@inproceedings{yin2022vit,
  title={A-vit: Adaptive tokens for efficient vision transformer},
  author={Yin, Hongxu and Vahdat, Arash and Alvarez, Jose M and Mallya, Arun and Kautz, Jan and Molchanov, Pavlo},
  booktitle={Proceedings of the IEEE/CVF conference on computer vision and pattern recognition},
  pages={10809--10818},
  year={2022}
}

@inproceedings{liu2024mmbench,
  title={Mmbench: Is your multi-modal model an all-around player?},
  author={Liu, Yuan and Duan, Haodong and Zhang, Yuanhan and Li, Bo and Zhang, Songyang and Zhao, Wangbo and Yuan, Yike and Wang, Jiaqi and He, Conghui and Liu, Ziwei and others},
  booktitle={European conference on computer vision},
  pages={216--233},
  year={2024},
  organization={Springer}
}

@article{chen2024we,
  title={Are we on the right way for evaluating large vision-language models?},
  author={Chen, Lin and Li, Jinsong and Dong, Xiaoyi and Zhang, Pan and Zang, Yuhang and Chen, Zehui and Duan, Haodong and Wang, Jiaqi and Qiao, Yu and Lin, Dahua and others},
  journal={Advances in Neural Information Processing Systems},
  volume={37},
  pages={27056--27087},
  year={2024}
}

@inproceedings{yue2024mmmu,
  title={Mmmu: A massive multi-discipline multimodal understanding and reasoning benchmark for expert agi},
  author={Yue, Xiang and Ni, Yuansheng and Zhang, Kai and Zheng, Tianyu and Liu, Ruoqi and Zhang, Ge and Stevens, Samuel and Jiang, Dongfu and Ren, Weiming and Sun, Yuxuan and others},
  booktitle={Proceedings of the IEEE/CVF conference on computer vision and pattern recognition},
  pages={9556--9567},
  year={2024}
}

@article{lu2023mathvista,
  title={Mathvista: Evaluating mathematical reasoning of foundation models in visual contexts},
  author={Lu, Pan and Bansal, Hritik and Xia, Tony and Liu, Jiacheng and Li, Chunyuan and Hajishirzi, Hannaneh and Cheng, Hao and Chang, Kai-Wei and Galley, Michel and Gao, Jianfeng},
  journal={arXiv preprint arXiv:2310.02255},
  year={2023}
}

@inproceedings{kembhavi2016diagram,
  title={A diagram is worth a dozen images},
  author={Kembhavi, Aniruddha and Salvato, Mike and Kolve, Eric and Seo, Minjoon and Hajishirzi, Hannaneh and Farhadi, Ali},
  booktitle={European conference on computer vision},
  pages={235--251},
  year={2016},
  organization={Springer}
}

@inproceedings{guan2024hallusionbench,
  title={Hallusionbench: an advanced diagnostic suite for entangled language hallucination and visual illusion in large vision-language models},
  author={Guan, Tianrui and Liu, Fuxiao and Wu, Xiyang and Xian, Ruiqi and Li, Zongxia and Liu, Xiaoyu and Wang, Xijun and Chen, Lichang and Huang, Furong and Yacoob, Yaser and others},
  booktitle={Proceedings of the IEEE/CVF conference on computer vision and pattern recognition},
  pages={14375--14385},
  year={2024}
}

@article{liu2024ocrbench,
  title={Ocrbench: on the hidden mystery of ocr in large multimodal models},
  author={Liu, Yuliang and Li, Zhang and Huang, Mingxin and Yang, Biao and Yu, Wenwen and Li, Chunyuan and Yin, Xu-Cheng and Liu, Cheng-Lin and Jin, Lianwen and Bai, Xiang},
  journal={Science China Information Sciences},
  volume={67},
  number={12},
  pages={220102},
  year={2024},
  publisher={Springer}
}

@article{yang2025qwen3,
  title={Qwen3 technical report},
  author={Yang, An and Li, Anfeng and Yang, Baosong and Zhang, Beichen and Hui, Binyuan and Zheng, Bo and Yu, Bowen and Gao, Chang and Huang, Chengen and Lv, Chenxu and others},
  journal={arXiv preprint arXiv:2505.09388},
  year={2025}
}

@article{team2025kimi,
  title={Kimi-vl technical report},
  author={Team, Kimi and Du, Angang and Yin, Bohong and Xing, Bowei and Qu, Bowen and Wang, Bowen and Chen, Cheng and Zhang, Chenlin and Du, Chenzhuang and Wei, Chu and others},
  journal={arXiv preprint arXiv:2504.07491},
  year={2025}
}
}


\newpage
\appendix
\setcounter{table}{0}
\setcounter{figure}{0}
\setcounter{equation}{0}
\renewcommand{\thetable}{A\arabic{table}}
\renewcommand{\thefigure}{A\arabic{figure}}
\renewcommand{\theequation}{A\arabic{equation}}

\section{Related Work}

\subsection{Vision Encoder}

Language-supervised contrastive models remain the dominant choice for MLLMs due to their natural pre-alignment with language. CLIP~\cite{radford2021learning} and its variants~\cite{zhai2023sigmoid, ilharco2021openclip, xu2023demystifying, sun2023eva} have progressively refined this paradigm through improved objectives, data curation and parameter scale. More recently, SigLIP 2~\cite{tschannen2025siglip} unifies contrastive, captioning, self-distillation and masked prediction objectives into a single recipe, achieving broad improvements in classification, localization, and MLLM transfer. Despite their dominance, these encoders inherit a language bottleneckk: they capture only what alt-text describes and exhibit "CLIP-blind" failures on fine-grained spatial distinctions, and most operate at fixed low resolutions. To push beyond these limits, a parallel line scales the visual backbone itself, exemplified by InternViT-6B~\cite{chen2024internvl} and AIMv2~\cite{fini2025multimodal}, while NaViT~\cite{dehghani2023patch} and the native-resolution ViTs of Qwen2-VL~\cite{wang2024qwen2}, Kimi K2.5~\cite{team2026kimi} make token count scale with image area. Another major line keeps the encoder fixed and instead partitions high-resolution inputs into multiple low-resolution slices that are encoded independently, as in LLaVA-NeXT~\cite{liu2024llavanext}, Intern VL 1.5~\cite{chen2024far}, LLaVA-UHD~\cite{guo2024llava} and mPLUG-DocOwl 1.5~\cite{hu2024mplug}. While effective at preserving fine-grained details with off-the-shelf encoders, slicing multiplies visual tokens and fragments cross-slice spatial context. However, scaling the encoder to billions of parameters or to native high resolutions inflates visual token counts and pretraining cost, creating a tension between visual fidelity and MLLMs efficiency.

\subsection{Multimodal Connector}

Bridging a vision encoder to an LLM requires a connector module, and the field has converged on two dominant designs. Query-based resamplers were popularized by Flamingo~\cite{alayrac2022flamingo}'s Perceiver Resampler, which compresses arbitrary spatio-temporal feature grids to a fixed 64 latent tokens via cross-attention with learned queries, and by BLIP-2~\cite{li2023blip}'s Q-Former, a 32-query bottleneck transformer pretrained with contrastive matching and generative objectives. This recipe was widely inherited: MiniGPT-4~\cite{zhu2023minigpt} freezes BLIP-2's Q-Former and trains only a linear head, InstructBLIP~\cite{dai2023instructblip} makes the queries instruction-aware, Qwen-VL~\cite{bai2023qwenvlversatilevisionlanguagemodel} employs a single-layer cross-attention compressor producing 256 tokens. Kosmos-1/2~\cite{huang2023language, peng2023kosmos} and mPLUG-Owl~\cite{ye2023mplug} all adopt Perceiver- or abstractor-style pooling, primarily for token-count efficiency. Projection-based connectors offer a competing minimalist alternative: LLaVA~\cite{liu2023visual}'s single linear layer and LLaVA-1.5~\cite{liu2024improved}'s two-layer GELU MLP retain every patch token, showing that simple token-preserving projection can match or exceed resamplers trained on orders of magnitude more data, and this design has since been widely adopted by many subsequent MLLMs~\cite{liu2024llavanext, chen2024sharegpt4v, li2025mini, karamcheti2024prismatic, wang2024cogvlm}. Yet the empirical record is contradictory: Honeybee~\cite{cha2024honeybee} attribute large gains to locality-preserving projection, whereas MM1~\cite{mckinzie2024mm1} finds connector architecture nearly negligible relative to image resolution and visual-token count. This leaves the trade-off between information fidelity and token efficiency unresolved and motivating a direct empirical comparison between Resampler- and MLP-style connectors.

\subsection{Token Compression}

The hundreds to thousands of visual tokens produced make token compression a central concern for MLLM efficiency. Existing approaches operate at three points of the pipeline. Inside the LLM, a line of largely training-free methods prunes visual tokens between transformer layers, exploiting the observation that visual tokens become increasingly redundant at deeper layers. FastV~\cite{chen2024image} drops low-attention visual tokens after an early layer, while SparseVLM~\cite{zhang2024sparsevlm}, VTW~\cite{lin2025boosting}, and PyramidDrop~\cite{xing2024pyramiddrop} extend this idea with text-aware or progressive schedules. Such methods are simple to deploy but inherit whatever redundancy the encoder has already produced. Between the encoder and the LLM, a learnable compressor distills patch tokens before they enter the language model. Inside the ViT, compression directly reduces the cost of the visual backbone itself. ToMe~\cite{bolya2022token} bipartite-matches and merges similar tokens at each layer without retraining; DynamicViT~\cite{rao2021dynamicvit} and A-ViT~\cite{yin2022vit} learn to drop uninformative tokens during the forward pass. In-encoder compression accelerates the entire backbone, but is tightly coupled to the encoder's pretraining objective and risks discarding tokens that downstream language grounding would have relied on.

\begin{table}[t!]
  \centering
  \caption{\textbf{Detailed results for the robustness study of slice-based encoding.} Detailed breakdown of Table~\ref{tab:encoding_robust} across the eight benchmarks, covering both the MoonViT backbone and the higher-resolution slicing schedule under compression rate $16\times$.}
  \footnotesize
  \label{app:encoding_robust_details}
  \setlength{\tabcolsep}{4pt}
  \resizebox{\linewidth}{!}{
  \begin{tabular}{l|ccccccccc|c}
    \toprule
    Data Scale & Method & MMMU & MathVista & MMB$_\text{EN}$ & MMB$_\text{CN}$ & MMStar & HallBench & AI2D & OCRBench & Avg. \\
    \midrule
    \rowcolor{blue!7} \multicolumn{11}{c}{MoonViT (Compression Rate 16$\times$)} \\
    \multirow{2}{*}{8M} & GE & 57.8 & 69.0 & \textbf{82.9} & 82.2 & 61.3 & 50.7 & 80.1 & 78.0 & 70.3 \\
     & SE & \textbf{58.8} & \textbf{70.1} & 82.7 & 82.2 & \textbf{64.4} & \textbf{52.0} & 80.1 & \textbf{82.2} & \textbf{71.6} \\
    \hline
    \multirow{2}{*}{16M} & GE & 57.7 & \textbf{73.4} & \textbf{83.8} & 82.6 & 65.3 & 53.3 & \textbf{82.7} & 79.0 & 72.2 \\
     & SE & \textbf{62.4} & 72.2 & 83.6 & \textbf{82.9} & \textbf{66.3} & \textbf{54.1} & 81.8 & \textbf{85.1} & \textbf{73.6} \\
    \rowcolor{blue!7} \multicolumn{11}{c}{Higher-Resolution (Compression Rate 16$\times$)} \\
    \multirow{2}{*}{8M} & GE & 56.4 & 66.2 & 82.6 & 82.0 & 61.1 & 48.4 & \textbf{79.7} & 73.9 & 68.8 \\
     & SE & \textbf{59.1} & \textbf{68.4} & \textbf{84.4} & \textbf{83.3} & \textbf{62.4} & \textbf{49.9} & 79.1 & \textbf{81.5} & \textbf{71.0} \\
    \bottomrule
  \end{tabular}}
\end{table}

\begin{table}[t!]
  \centering
  \caption{\textbf{Main comparison on in-house data across training scales.} Both systems share an identical architecture, training recipe, data, and end-to-end $16\times$ compression ratio; they differ only in where compression occurs. Avg.\ is computed over the eight benchmarks shown. \textit{Post-ViT baseline} performs all compression after the ViT. \textit{Ours} performs $4\times$ compression inside the ViT after layer 6 and another $4\times$ after the ViT.}
  \footnotesize
  \setlength{\tabcolsep}{4pt}
  \label{tab:main_inhouse_full}
  \begin{tabular}{l|c|cccccccc|c}
    \toprule
    Data Scale & Method & MMMU & MathVista & MMB$_\text{EN}$ & MMB$_\text{CN}$ & MMStar & HallBench & AI2D & OCRBench & Avg. \\
    \midrule
    \multirow{2}{*}{4M}  & Post-ViT & 57.9 & \textbf{63.0} & \textbf{79.4} & \textbf{79.1} & \textbf{60.6} & \textbf{50.5} & \textbf{77.7} & \textbf{77.5} & \textbf{68.2} \\
                         & Ours      & \textbf{60.3} & 61.7 & 78.6 & 78.4 & 60.4 & 47.7 & 76.6 & 75.3 & 67.4 \\
    \hline
    \multirow{2}{*}{8M}  & Post-ViT & 58.6 & 67.3 & \textbf{83.7} & \textbf{82.3} & 62.9 & 51.2 & 79.8 & \textbf{79.1} & 70.6 \\
                         & Ours      & \textbf{59.6} & \textbf{68.6} & 83.4 & 81.6 & 62.9 & \textbf{52.0} & \textbf{80.6} & 76.7 & \textbf{70.7} \\
    \hline
    \multirow{2}{*}{16M} & Post-ViT & 59.1 & 71.0 & \textbf{84.9} & \textbf{83.5} & \textbf{65.5} & 51.5 & 81.2 & 83.2 & 72.5 \\
                         & Ours      & \textbf{61.2} & \textbf{71.1} & 84.1 & 83.3 & 65.3 & \textbf{54.7} & \textbf{81.8} & \textbf{83.5} & \textbf{73.1} \\
    \hline
    \multirow{2}{*}{32M} & Post-ViT & \textbf{63.6} & \textbf{72.7} & \textbf{85.5} & 84.9 & 65.9 & \textbf{53.6} & \textbf{82.5} & \textbf{84.8} & \textbf{74.2} \\
                         & Ours      & 62.3 & 72.0 & 84.7 & \textbf{85.0} & \textbf{66.2} & 52.8 & 82.4 & 82.7 & 73.5 \\
    \hline
    \multirow{2}{*}{64M} & Post-ViT & \textbf{63.9} & 76.3 & \textbf{87.0} & 86.4 & \textbf{67.9} & \textbf{56.5} & 84.7 & \textbf{86.7} & \textbf{76.2} \\
                         & Ours      & 61.9 & \textbf{76.9} & 86.2 & \textbf{86.5} & 66.9 & 55.2 & \textbf{84.9} & 85.9 & 75.6 \\
    \bottomrule
  \end{tabular}
\end{table}

\section{Detailed Analysis and Results}

\subsection{Detailed Analysis of Encoding Strategies}
\label{app:se_ge_analysis}

Across the evaluated SigLIP 2 settings, MoonViT settings, and slicing schedules, slice-based encoding (SE) improves the average score over global encoding (GE), although individual benchmark outcomes remain mixed. The MoonViT comparison shows that this average advantage persists even with a backbone designed for native-resolution processing, and the higher-resolution slicing variant further suggests that the result is not tied to a single slicing budget. We therefore interpret SE not merely as a computational workaround, but as an encoding strategy that changes the context in which visual features are formed before compression.

The key difference lies less in the compression ratio itself than in the attention context used by the ViT. With the pixel-unshuffle MLP compressor, both GE and SE apply a locality-preserving spatial merge, so the compressor does not globally pool all visual tokens. However, the features entering this compressor have been produced under different encoding contexts. GE encodes the full image in a single ViT forward pass, where all patches interact in one global attention space. SE decomposes the image into a thumbnail and spatially coherent slices, then encodes each slice independently, so the ViT forms features within localized views before those features are spatially merged.

This local encoding bias is especially relevant for fine-grained perception. GE preserves unrestricted patch-to-patch interaction inside the ViT, which is useful for global context but may dilute the inductive bias toward local structure. SE sacrifices some within-ViT global interaction, yet it encourages the visual encoder to extract text, chart marks, and dense document patterns within local neighborhoods before the same type of spatial compression is applied. The largest and most stable gains on OCRBench are consistent with this interpretation: tasks that depend heavily on small local structures appear to benefit from forming visual features in localized views before compression. Within our tested settings, the advantage of SE therefore appears to come less from the compressor itself and more from the locality of the preceding visual encoding.

\begin{table}[t!]
  \centering
  \caption{\textbf{Full per-benchmark results for in-ViT compression design ablations.} All variants share the same end-to-end $16\times$ compression ratio and insertion depth $k=6$, differing only in how the $4\times$ in-ViT compression stage is realized. FLOPs are reported per slice through the ViT, and bold marks the best score in each column.}
  \footnotesize
  \label{app:ablation_full}
  \setlength{\tabcolsep}{4pt}
  \resizebox{\linewidth}{!}{
  \begin{tabular}{l|c|cccccccc|c}
  \toprule
  Method & FLOPs (G) & MMMU & MathVista & MMB$_\text{EN}$ & MMB$_\text{CN}$ & MMStar & HallBench & AI2D & OCRBench & Avg. \\
  \midrule
  \rowcolor{blue!7} \multicolumn{11}{c}{\textit{Post-ViT merging}} \\
  Post-ViT Baseline 
  & 3555.1 & 58.6 & 67.3 & \textbf{83.7} & \textbf{82.3} & \textbf{62.9} & 51.2 & 79.8 & 79.1 & 70.6 \\
  
  \midrule
  \rowcolor{blue!7} \multicolumn{11}{c}{\textit{Naive in-ViT merging}} \\
  Average Pool 
  & \textbf{1368.7} & 59.2 & 67.2 & 83.6 & 81.5 & 62.4 & 47.1 & 79.8 & 75.7 & 69.6 \\
  Pixel-Unshuffle MLP 
  & 1401.2 & 58.7 & 66.7 & 82.4 & 81.4 & 61.6 & 49.2 & 80.0 & 78.6 & 69.8 \\
  Reused MLP 
  & 1490.2 & 57.6 & 67.0 & 81.8 & 81.3 & 62.3 & 48.8 & \textbf{81.0} & \textbf{79.5} & 69.9 \\
  
  \midrule
  \rowcolor{blue!7} \multicolumn{11}{c}{\textit{Cross-attention merging}} \\
  Cross-Attn (top-left query)
  & 1402.0 & 59.9 & \textbf{68.6} & 83.6 & 81.5 & 61.1 & 50.8 & 80.1 & 78.2 & 70.5 \\
  Cross-Attn (mean query)
  & 1402.0 & \textbf{61.0} & 66.0 & 82.2 & 81.5 & 61.5 & 47.5 & 80.6 & 78.5 & 69.9 \\
  
  \midrule
  \rowcolor{blue!7} \multicolumn{11}{c}{\textit{Window-attention merging}} \\
  Win-Attn w/ MLP 
  & 1484.1 & 58.8 & 67.4 & 83.5 & 81.7 & 62.7 & 47.3 & 80.5 & 78.9 & 70.1 \\
  Win-Attn w/ Reused MLP 
  & 1573.1 & 59.6 & \textbf{68.6} & 83.4 & 81.6 & \textbf{62.9} & \textbf{52.0} & 80.6 & 76.7 & \textbf{70.7} \\
  \bottomrule
  \end{tabular}}
\end{table}

\subsection{Detailed Results of Connector Designs}
\label{app:connector_analysis}

\begin{table}[t!]
  \centering
  \caption{\textbf{Comparison of connector designs.}
  We compare the MLP downsampler against the resampler under the SE setting across multiple downsampling rates. OCRBench is divided by 10, and Avg.\ is computed over the eight benchmarks shown.}
  \footnotesize
  \setlength{\tabcolsep}{4pt}
  \label{tab:connector_detailed}
  \resizebox{\linewidth}{!}{
  \begin{tabular}{l|l|cccccccc|c}
    \toprule
    Data Scale & Connector & MMMU & MathVista & MMB$_\text{EN}$ & MMB$_\text{CN}$ & MMStar & HallBench & AI2D & OCRBench & Avg. \\
    \midrule
    \rowcolor{blue!7}
    \multicolumn{11}{c}{Downsampling Rate 4$\times$} \\
    \midrule
    \multirow{2}{*}{4M}
      & Resampler & 57.4 & 62.7 & 80.3 & 78.9 & 60.7 & 46.2 & 78.1 & 73.9 & 67.3 \\
      & MLP       & \textbf{61.9} & \textbf{66.7} & \textbf{82.9} & \textbf{79.5} & \textbf{62.3} & \textbf{49.1} & \textbf{80.5} & \textbf{82.0} & \textbf{70.6} \\
    \cmidrule(lr){1-11}
    \multirow{2}{*}{8M}
      & Resampler & 57.9 & 61.7 & 80.4 & 77.9 & 58.9 & 49.1 & 78.2 & 68.7 & 66.6 \\
      & MLP       & \textbf{60.3} & \textbf{71.2} & \textbf{85.2} & \textbf{83.4} & \textbf{64.3} & \textbf{56.3} & \textbf{82.0} & \textbf{83.6} & \textbf{73.3} \\
    \midrule
    \rowcolor{blue!7}
    \multicolumn{11}{c}{Downsampling Rate 16$\times$} \\
    \midrule
    \multirow{2}{*}{4M}
      & Resampler & \textbf{58.7} & 62.3 & \textbf{79.6} & 78.2 & 59.7 & 49.5 & 76.9 & 75.1 & 67.5 \\
      & MLP       & 57.9 & \textbf{63.0} & 79.4 & \textbf{79.1} & \textbf{60.6} & \textbf{50.5} & \textbf{77.7} & \textbf{77.5} & \textbf{68.2} \\
    \cmidrule(lr){1-11}
    \multirow{2}{*}{8M}
      & Resampler & 57.1 & 65.9 & 81.9 & 81.3 & 61.3 & 49.1 & \textbf{80.3} & 78.3 & 69.4 \\
      & MLP       & \textbf{58.6} & \textbf{67.3} & \textbf{83.7} & \textbf{82.3} & \textbf{62.9} & \textbf{51.2} & 79.8 & \textbf{79.1} & \textbf{70.6} \\
    \cmidrule(lr){1-11}
    \multirow{2}{*}{16M}
      & Resampler & 59.1 & 69.1 & 84.0 & 83.5 & 64.1 & \textbf{54.3} & 81.2 & 81.2 & 72.1 \\
      & MLP       & 59.1 & \textbf{71.0} & \textbf{84.9} & 83.5 & \textbf{65.5} & 51.5 & 81.2 & \textbf{83.2} & \textbf{72.5} \\
    \bottomrule
  \end{tabular}}
\end{table}

The detailed results in Table~\ref{tab:connector_detailed} clarify why we use the MLP connector as the post-ViT baseline. Its largest gains appear at $4\times$ compression, where the output sequence still preserves a relatively rich coarse layout. In this regime, pixel-unshuffle can exploit its built-in spatial structure: each output token is formed from a fixed local patch group and remains tied to a local image neighborhood. The resampler, by contrast, summarizes the ViT output through learnable queries, so its outputs no longer have fixed spatial correspondence and must learn this organization from data.

As compression becomes more aggressive, the gap narrows but does not reverse. At $16\times$ compression, both connectors must discard more spatial detail, reducing the benefit of an explicitly locality-preserving merge. Even in the most favorable setting for the resampler, with 16M training samples, MLP remains slightly ahead. This suggests that the resampler can partially learn useful aggregation with enough data and a tight token budget, but it does not provide a stronger default than the simpler spatially structured connector. We therefore use the MLP connector as the strongest post-ViT baseline before asking whether part of the compression should be moved inside the ViT.

\subsection{Additional Ablations on the Open-Source LLaVA-OneVision Setting}

\begin{table}[t!]
  \centering
  \caption{\textbf{Ablation on the open-source LLaVA-OneVision setting.} We evaluate different in-ViT compressor designs under the open-source dataset.}
  \footnotesize
  \label{tab:ablation_onevision}
  \setlength{\tabcolsep}{4pt}
  \resizebox{\linewidth}{!}{
  \begin{tabular}{lcccccccc|c}
    \toprule
    Method & MMMU & MathVista & MMB$_\text{EN}$ & MMB$_\text{CN}$ & MMStar & HallBench & AI2D & OCRBench & Avg. \\
    \midrule
    \rowcolor{blue!7} \multicolumn{10}{c}{LLaVA-OneVision Open-source Setting} \\
    Post-ViT Baseline & 46.3 & 62.2 & 74.9 & 71.6 & 56.7 & 40.3 & 79.9 & 64.7 & 62.1 \\
    Average Pool & 47.6 & 62.4 & 75.4 & 73.1 & 56.3 & 40.3 & 81.5 & 62.9 & 62.4 \\
    Pixel-Unshuffle MLP & 46.6 & 62.3 & 72.8 & 72.2 & 51.7 & 38.3 & 80.2 & 58.7 & 60.4 \\
    Reused MLP & 45.3 & 60.4 & 76.1 & 74.1 & 55.3 & 40.7 & 81.2 & 63.7 & 62.1 \\
    Cross-Attn (top-left) & 48.6 & 62.0 & 75.1 & 72.5 & 56.4 & \textbf{44.7} & 80.5 & 64.8 & 63.1 \\
    Cross-Attn (mean) & 47.6 & 62.4 & 75.4 & 73.1 & 56.3 & 40.3 & 81.5 & 62.9 & 62.4 \\
    Win-Attn w/ MLP & \textbf{50.9} & 61.4 & 75.4 & \textbf{73.9} & 54.7 & 42.7 & \textbf{81.8} & \textbf{65.0} & 63.2 \\
    Win-Attn w/ Reused MLP & 48.3 & \textbf{63.5} & \textbf{76.7} & 73.5 & \textbf{57.0} & 42.7 & 81.1 & 64.6 & \textbf{63.4} \\
    \bottomrule
  \end{tabular}}
\end{table}

Table~\ref{tab:ablation_onevision} further evaluates the same family of in-ViT downsampling designs under the open-source LLaVA-OneVision training setting. The trend is broadly consistent with the in-house ablations in the main paper: naively inserting a learnable MLP merger inside the ViT is not sufficient, as the plain MLP variant drops from the baseline average of $62.1$ to $60.4$. In contrast, designs that introduce local interaction before token reduction are substantially more robust. Cross-attention and window-attention variants improve over the plain MLP, suggesting that early compression benefits from first allowing the tokens within each local $2\times2$ region to exchange information.

Among all variants, \textit{Win-Attn w/ Reused MLP} achieves the best average score, improving the baseline from $62.1$ to $63.4$. The gain is modest but consistent with the main-paper conclusion: local contextualization and parameter-reuse initialization are complementary. Compared with \textit{Win-Attn w/ MLP}, reuse improves the average score from $63.2$ to $63.4$. This mixed per-benchmark pattern indicates that the open-source setting is somewhat noisier, but the best average performance still comes from the reused window-attention design, supporting its transfer beyond the in-house training recipe.

\begin{table}[t!]
  \centering
  \caption{\textbf{Comparison of different ViT internal downsampling strategies across training scales.} All systems share an identical architecture, training recipe, data, and end-to-end $16\times$ compression ratio. They differ only in the downsampling module design.}
  \footnotesize
  \setlength{\tabcolsep}{4pt}
  \label{tab:ablation_scale_inhouse}
  \resizebox{\linewidth}{!}{
  \begin{tabular}{l|c|cccccccc|c}
    \toprule
    Data Scale & Method & MMMU & MathVista & MMB$_\text{EN}$ & MMB$_\text{CN}$ & MMStar & HallBench & AI2D & OCRBench & Avg. \\
    \midrule
    \multirow{3}{*}{8M}  & Win-Attn w/ Reused MLP & 59.6 & \textbf{68.6} & 83.4 & \textbf{81.6} & \textbf{62.9} & \textbf{52.0} & \textbf{80.6} & 76.7 & \textbf{70.7} \\
                         & Cross-Attn (top-left)  & 59.9 & \textbf{68.6} & \textbf{83.6} & 81.5 & 61.1 & 50.8 & 80.1 & 78.2 & 70.5 \\
                         & Cross-Attn (mean)      & \textbf{61.0} & 66.0 & 82.2 & 81.5 & 61.5 & 47.5 & \textbf{80.6} & \textbf{78.5} & 69.8 \\
    \hline
    \multirow{3}{*}{16M} & Win-Attn w/ Reused MLP & \textbf{61.2} & \textbf{71.1} & 84.1 & \textbf{83.7} & \textbf{65.3} & 54.7 & 81.8 & \textbf{83.5} & \textbf{73.1} \\
                         & Cross-Attn (top-left)  & \textbf{61.2} & 69.2 & \textbf{85.2} & \textbf{83.7} & 63.5 & 52.6 & \textbf{82.3} & 81.0 & 72.3 \\
                         & Cross-Attn (mean)      & 61.0 & 69.3 & 84.6 & 83.1 & 64.4 & \textbf{55.3} & 81.4 & 83.2 & 72.8 \\
    \bottomrule
  \end{tabular}}
\end{table}

\section{Hyperparameters}
\label{app:hparams}
Table~\ref{tab:tunnel_training_hparams} and Table~\ref{tab:tunnel_llava_rebalanced_hparams} provide the detailed optimization settings for the four-stage training recipe described in Section~\ref{sec:exp_setup}. Both recipes begin with a warmup stage for vision-language alignment, continue with high-quality image training, and end with supervised instruction tuning. The tables report the learning-rate schedule, training length, warmup steps, trainable modules, and packing-equivalent per-GPU batch size for the in-house data setting and the LLaVA-OneVision training setting, respectively.

\begin{table}[t]
  \centering
  \small
  \caption{Training hyperparameters on in-house data.}
  \label{tab:tunnel_training_hparams}
  \begin{tabular}{lcccccc}
  \toprule
  Stage & LR & LR$_{\min}$
  & Trainable & Batch size \\
  \midrule
  1
  & $1.0{\times}10^{-4}$ & $5.0{\times}10^{-5}$
  & ViT / Connector
  & 32 \\
  
  2
  & $1.0{\times}10^{-5}$ & $5.0{\times}10^{-6}$
  & ViT
  & 6 \\
  
  3
  & $5.0{\times}10^{-5}$ & $1.0{\times}10^{-5}$
  & Full
  & 6 \\
  
  4
  & $1.0{\times}10^{-5}$ & $1.0{\times}10^{-6}$
  & Full
  & 9 \\
  \bottomrule
  \end{tabular}
  \end{table}

\begin{table}[t!]
\centering
\small
\caption{Training hyperparameters on LLaVA-OneVision data.}
\label{tab:tunnel_llava_rebalanced_hparams}
\begin{tabular}{lcccccc}
\toprule
Stage & LR & LR$_{\min}$ 
& Trainable & Batch size \\
\midrule
1
& $1.0{\times}10^{-4}$ & $5.0{\times}10^{-5}$
& ViT / Connector
& 16 \\

2
& $1.0{\times}10^{-5}$ & $5.0{\times}10^{-6}$
& Full
& 20 \\

3
& $5.0{\times}10^{-5}$ & $1.0{\times}10^{-5}$
& Full
& 34 \\

4
& $1.0{\times}10^{-5}$ & $1.0{\times}10^{-6}$
& Full
& 11 \\
\bottomrule
\end{tabular}

\end{table}

\section{Limitations}
\label{app:limitations}

While LLaVA-UHD v4 significantly accelerates high-resolution visual encoding, several limitations remain for future work. First, our intra-ViT compression module applies a fixed and uniform spatial downsampling rate across all patches. It does not adapt to the varying information density within an image, making dynamic, content-aware token reduction (e.g., allocating more tokens to dense text and fewer to plain backgrounds) an important next step. Second, the optimal insertion depth for the compressor ($k=6$) was empirically determined for the SigLIP 2 backbone; migrating to architecturally distinct or substantially deeper vision encoders may require re-evaluating this hyperparameter. Finally, although slice-based encoding excels at fine-grained perception, it inherently fragments high-resolution context across slice boundaries, relying primarily on the low-resolution thumbnail to bridge global interactions.

\end{document}